\documentclass[review]{elsarticle}

\usepackage[a4paper, total={6in, 9in}]{geometry}

\usepackage{lineno,hyperref}
\modulolinenumbers[5]

\usepackage{setspace}
\usepackage[utf8]{inputenc}
\usepackage{times}
\usepackage{graphicx}
\usepackage[dvipsnames]{xcolor}
\usepackage{tikz}
\usepackage{comment}
\usepackage{amsmath,amssymb} 
\usepackage{float}
\usepackage{multirow}
\usepackage{adjustbox}
\usepackage{booktabs}
\usepackage{url}
\usepackage{hyperref}
\usepackage{tabularx}
\usepackage{xcolor}

\usepackage{algpseudocode}
\usepackage[ruled,lined,linesnumbered,commentsnumbered,longend]{algorithm2e}

\DeclareGraphicsExtensions{.pdf,.png,.jpg}

\begin{document}


\begin{frontmatter}

\title{Global Point Cloud Registration Network for Large Transformations}

\author[1]{Hanz Cuevas-Velasquez}
\ead{hanz.cuevas@tuebingen.mpg.de}

\author[2]{Alejandro Galán-Cuenca\corref{corresponding}}
\cortext[corresponding]{Corresponding author: 
Email: a.galan@ua.es}
\ead{a.galan@ua.es}

\author[2]{Antonio Javier Gallego}
\ead{jgallego@dlsi.ua.es}

\author[2]{Marcelo Saval-Calvo}
\ead{m.saval@ua.es}

\author[3]{Robert B. Fisher}
\ead{rbf@inf.ed.ac.uk}

\address[1]{Max Planck Institute for Intelligent Systems, 72076, Max Planck Ring 4, Tuebingen, Germany}

\address[2]{University Institute for Computing Research, University of Alicante, E-03690 Alicante, Spain}

\address[3]{School of Informatics, University of Edinburgh, Edinburgh, EH8 9AB, UK}

\begin{abstract}
Three-dimensional data registration is an established yet challenging problem that is key in many different applications, such as mapping the environment for autonomous vehicles, and modeling objects and people for avatar creation, among many others. Registration refers to the process of mapping multiple data into the same coordinate system by means of matching correspondences and transformation estimation. Novel proposals exploit the benefits of deep learning architectures for this purpose, as they learn the best features for the data, providing better matches and hence results. However, the state of the art is usually focused on cases of relatively small transformations, although in certain applications and in a real and practical environment, large transformations are very common. In this paper, we present \textit{ReLaTo} (Registration for Large Transformations), an architecture that faces the cases where large transformations happen while maintaining good performance for local transformations. This proposal uses a novel Softmax pooling layer to find correspondences in a bilateral consensus manner between two point sets, sampling the most confident matches. These matches are used to estimate a coarse and global registration using weighted Singular Value Decomposition (SVD). A target-guided denoising step is then applied to both the obtained matches and latent features, estimating the final fine registration considering the local geometry. All these steps are carried out following an end-to-end approach, which has been shown to improve 10 state-of-the-art registration methods in two datasets commonly used for this task (ModelNet40 and KITTI), especially in the case of large transformations.
\end{abstract}


\end{frontmatter}


\section{Introduction}

Registration is a major problem in computer vision, medical imaging, remote sensing, robotics, and photogrammetry. It refers to the process of transforming different data sets to align them in the same coordinate system by applying operations to map the data either spatially or with respect to other criteria, such as intensity in the case of photograph registration for panorama creation. In this paper, we will focus on the case of three-dimensional Point Cloud Registration (PCR), which is involved in multiple applications that require finding the pose transformation when the observer or sensor changes the viewpoint. Some examples of these applications are model reconstruction from multi-view data~\cite{Saval-Calvo2015}, where the surface shape is reconstructed by aligning multiple views, or in Simultaneous Localization and Mapping (SLAM)~\cite{Kim2018}, used in robot navigation and autonomous driving, which employs registration to consecutively merge the estimated map with new views. PCR can also be used in non-rigid (i.e., deformable) alignments, such as for body movement registration or in face expression modeling~\cite{Chaudhury2020,Saval-Calvo2018}.

\begin{figure}[!ht]
    \centering 
    \includegraphics[width=\textwidth]{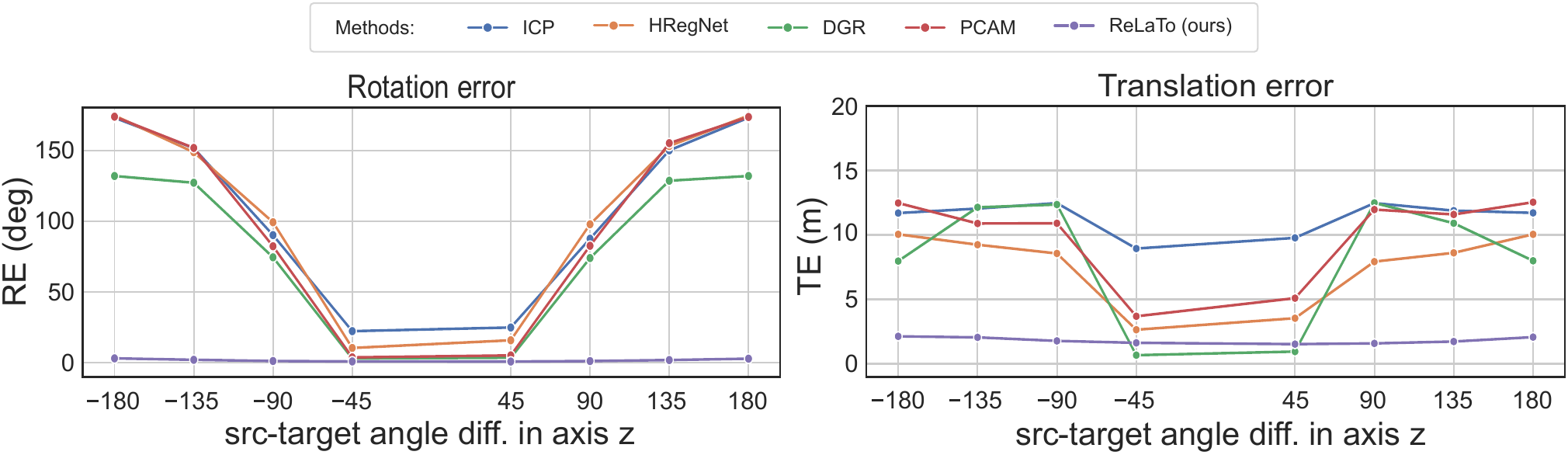}
    \caption{Comparison between the error made by the proposed model and the state of the art under large transformations (considering rotations on the $z$ axis) for the KITTI dataset~\cite{Menze2015}. The first image shows the rotation error (RE) and the second is the translation error (TE). Note that the proposed method results in almost no RE. For TE, although one state-of-the-art solution marginally outperforms the proposed model for transformations smaller than 45 degrees, our approach demonstrates superior average performance across the entire range of transformations.}
    \label{fig:kitti_rotation_errors}
\end{figure}

PCR has been extensively studied for several different applications, resulting in a large body of related literature~\cite{pomerleau2015,Villena-Martinez2020}. However, most research has focused on the ``fine'' registration---i.e., when the transformation to be applied is less than 45 degrees. This can be seen in Figure~\ref{fig:kitti_rotation_errors}, where the results obtained by four representative state-of-the-art methods worsened notably when the transformation applied exceeds 45 degrees (with the exception of our proposal presented here). It is important to note that in a practical case and in many real applications, the scanned point clouds can often undergo large transformations~\cite{kaljaca2019,pu2018} either due to the use of cheap or low-frequency sensors---because of device constraints such as in embedded vision applications on devices with limited resources---or simply due to the speed of movement or rotation of the device that performs the capture. This, together with sparse and noisy measurements and incomplete observations, makes PCR---and more especially, the case of large transformations---a very challenging task of great interest, both on a practical and scientific level.

A general categorization of registration methods divides them into either \textit{global} and \textit{local}, or into \textit{coarse} and \textit{fine} approaches. Global registration refers to methods based on global features---those that consider the complete point cloud---which allow large rotations and/or translations, as opposed to those based on local features, in which the space of potential correspondences is restricted to the local environment. A distinction is also made between coarse and fine methods, in which an approximate (or coarse) alignment is first calculated and used as initialization for a more precise (or fine) second step. As will be detailed in the following state-of-the-art review section, each of these approaches has its advantages and disadvantages. Local and fine registration algorithms tend to fail if they are not provided with pre-aligned data, being necessary to previously apply global or coarse registration steps~\cite{Zhao2021,Villena-Martinez2021,Yuan2022,Yu2021}. In controlled environments, global registration can be achieved by calibrating the system, i.e., using predefined features to estimate the location of the different sensors with respect to a common reference system. However, this is not practical in most real scenarios, as the acquisition conditions may change. For instance, in SLAM, the vehicle takes different trajectories or faces new environments for which the camera setup cannot be pre-calibrated. 

Regardless of the level of registration to be carried out, this process is usually composed of two main parts: the correspondence matching between the point sets and the aligning transformation estimation. The initial stage establishes correspondences between the input sets, being critical for the accuracy of the final alignment. Many different approaches have been proposed to find good descriptors for match-finding, from traditional handcrafted descriptors~\cite{Guo2014,Yang2016} to more recent learning-based solutions~\cite{Charles2017,Yuan2020,Ao2021}. The matching is a critical step since the transformation estimation fully depends on it. There exist different approaches to improve the matches, such as the traditional RANSAC~\cite{Fischler1981} that rejects non-coherent correspondences, but fails when the number of correct matches is low; or the use of non-binary matches~\cite{myronenko2010}, where soft correspondences relaxes the matching estimation through one-to-many relationships. 

A key contribution of this paper is the introduction of a bilateral consensus for feature matching (explained in Section~\ref{sec:maxPool}), designed to secure optimal matches between point clouds in both source-to-target and vice versa directions.

The second stage of the registration process uses the matches to estimate the transformation to align one set with the other by applying different optimization techniques. In rigid registration, it is common to use Procrustes analysis to find the rotation matrix by means of Singular Value Decomposition (SVD)~\cite{Eggert1997}. This procedure can be enhanced through iterating until convergence, albeit at the cost of increased processing time. 

Another key contribution of our paper is a novel approach that incorporates a global-to-local strategy coupled with a ``target-guided denoising'' method to refine registration, which enhances the outcomes of the second stage of the registration process (refer to Section\ref{sec:refinement}).

Recent approaches to rigid registration incorporate learning-based techniques either in both parts of the pipeline or as an end-to-end solution that finds the transformation directly from the original data. A review can be found in~\cite{Villena-Martinez2020}, which concludes that most proposals do not address the coarse alignment problem with end-to-end architectures, but with two-step methods that include feature extraction and matching followed by transformation estimation~\cite{choy2020,Cao2021,Lu2021}. Section~\ref{sec:soa} provides a more complete review of the state of the art.

To fill the gap of solutions in the challenging problem of 3D rigid PCR in scenarios with noisy and incomplete data that require large transformations, we present \textit{ReLaTo} (Registration for Large Transformations) (Sec.~\ref{sec:proposal}). This end-to-end network architecture initially estimates local-global latent features from both sets of points to be registered (Sec.~\ref{sec:network}). It then searches for the best matches using a novel bilateral consensus criterion and Softmax pooling, facilitating the calculation of an initial global registration (outlined in Section~\ref{sec:maxPool}). 
The process ends with the application of a target-guided denoising technique to estimate the finer local transformation (see Section~\ref{sec:refinement}). Notably, ReLaTo learns the correspondences in an unsupervised way, relying solely on the confidence scores and the pairs selected by the proposed Softmax pooling layer. As has been shown in Figure~\ref{fig:kitti_rotation_errors}, our network attains good results in large transformation scenarios where other methods exhibit high error figures or even fail to converge. The efficacy and impact of the proposal will be further validated through an extensive set of experiments in Section~\ref{sec:experiments} and thoroughly discussed in Section~\ref{sec:results:ablation}.

In summary, this paper \textbf{makes the following contributions}: 

\begin{enumerate}
    \item A method for 3D point cloud coarse-to-fine global registration that is able to handle large-magnitude transformations. 
    
    \item A novel bilateral consensus estimation criteria for correspondence estimation. 
    
    \item A Softmax pooling layer that learns in an unsupervised manner the confidence score of pairs of points and samples the best ones. 
    
    \item A ``target-guided denoising'' layer that re-positions the source points based on the learned features to improve the final registration.
    
    \item An exhaustive comparison with different datasets and state-of-the-art methods that shows a remarkable superiority of the proposed method, especially for cases of large transformations.
\end{enumerate}


\section{Related work} 
\label{sec:soa}

As described previously, PCR relies on identifying good matches between points within source and target datasets, coupled with an accurate optimization algorithm to find the local minima. For this, the PCR pipeline is usually divided into two principal phases: 1) feature extraction and matching, where unique characteristics are identified to align the two point clouds, and 2) the estimation of the aligning transformation, which precisely determines how the point clouds should be positioned relative to each other. Features can be categorized as dense or scattered. Dense features, such as those utilized in methods like Iterative Closest Point (ICP)~\cite{besl1992} and Deep Weighted Consensus (DWC)~\cite{Ginzburg2022}, leverage all available data---3D points---along with their geometrical information and color/orientation. In contrast, scattered features offer a computationally efficient and robust alternative. Traditionally, handcrafted features~\cite{Han2023} were prevalent, but recent advancements have seen a shift towards learning-based methods~\cite{Zhang2022}, which demonstrate improved performance. Additionally, some approaches employ hierarchical feature extraction, enabling the algorithm to perform matching in a local-to-global manner~\cite{Lu2021}. Following the matching of features, incorrect matches are filtered out, and the transformation is optimized. Recent works are moving towards end-to-end solutions that simultaneously estimate features and transformations in a single step~\cite{Qian2023}.

The following sections delve into the works based on the two-step pipeline---extraction, matching, and alignment of features---and those proposing end-to-end architectures. In all cases, a brief overview of the classical methods is provided before moving on to the current trend based on deep learning.

\subsection{Feature-based matching}

The two-step pipeline for PCR has been extensively studied in classical methodologies. ICP, as introduced by Besl and McKay~\cite{besl1992}, stands out as one of the best-known algorithms for PCR. It employs the point coordinates---optionally incorporating color and orientation---to estimate the correspondences via nearest-neighbor, assuming an initial good alignment of the point sets, and solving a least-squares problem for finding the transformation. Several methods have been proposed to improve the speed and robustness of ICP~\cite{agamennoni2016, bouaziz2013, pomerleau2015}. Other classical approaches~\cite{myronenko2010,luo2001,pu2018} have focused on generalizing the binary ICP by adopting a soft-assignment strategy, which assigns matching probabilities across all point combinations. Although these proposals improve the original version of ICP, they often suffer from high computational complexity.

Instead of using point coordinates to find the correspondences, other approaches rely on local and global descriptors obtained by analyzing the 3D geometry of the point sets. Classical methods are usually based on handcrafted features~\cite{Tombari2010,Drost2012}, with Fast Point Feature Histograms (FPFH)~\cite{Rusu2009} as one of the most representative. More recently, these handcrafted features have been replaced by descriptors extracted from pre-trained neural networks~\cite{Charles2017,Qi2017,deng2018,Deng2018a,choy2019,Ao2021, Bai2021}, marking a significant advancement in the field. These approaches typically employ SVD to estimate the transformation based on the identified feature correspondences. A notable example is Deep Global Registration (DGR)~\cite{choy2020}, which uses features extracted from a pre-trained neural network and estimates the transformation using weighted SVD.

Since features are affected by noise, it is essential to remove erroneous matches (outliers) before proceeding with the transformation calculation. RANSAC~\cite{Fischler1981, Wei2023} stands out as the most recognized technique for this purpose~\cite{Jiang2022, Zhang2022}. Beyond RANSAC, other proposals have been proposed for outlier removal. For instance, PointDSC~\cite{Bai2021} enhances the identification of inlier matches by discarding outliers based on spatial consistency. Similarly, 3DRegNet~\cite{Pais2020} introduces a classification block specifically designed to assess correspondences as inliers or outliers, further refining the process of accurate match selection.

\subsection{End-to-end approaches} 

End-to-end methodologies utilize neural networks to simultaneously identify correspondences and compute the transformation. Deep Closest Point (DCP)~\cite{Wang2019a} was one of the first works that proposed an end-to-end network to solve PCR. It employs an attention matrix to identify point correspondences and utilizes SVD for transformation estimation. PRNet\cite{Wang2019b} further developed this concept, presenting an architecture that improves partial-to-partial registrations. Following this progression, HRegNet~\cite{Lu2021} proposed a network divided into two blocks: the first focuses on minimizing the distance between descriptors of matching pairs, while the second executes a coarse-to-fine registration leveraging the extracted features. PCAM~\cite{Cao2021} proposed a similar solution but identifying the matching points and confidence scores across all network layers, and also focused on minimizing the matching error instead of using SVD to solve the registration problem.

Recent advancements in PCR have focused on improving the confidence of matched pairs~\cite{Yu2021, Bai2021}, typically through optimizing a loss function to learn binary correspondence scores---where 1 represents inliners and 0 outliers. This implies that the ground truth has to label all correspondences even when there is noise or incomplete information. To mitigate this challenge, some strategies suggest labeling only pairs with distances below a specific threshold. While this approach is pragmatic, the inherent noise in real data complicates the selection of a suitable distance threshold. Among these, we find the RGM architecture~\cite{Fu2021, Fu2022}, which focuses on learning good correspondences to later apply SVD. This method exploits geometric information to estimate a graph-based feature space to find correspondences, where nodes encode local information, rather than using the actual point cloud. However, this may lead to incorrect registration if two different elements are matched, i.e., two views of an object with partial overlap, since there is no global information that may serve as a general reference. Other methods use graphs to extract characteristics from the point clouds and establish reliable matches~\cite{Zanfir2018}, such as Predator~\cite{Huang2021} and RRGA-Net~\cite{Qian2023}. RPM-Net~\cite{Yew2020}, inspired by the classic RPM~\cite{Yang2011}, adopts soft-assigned correspondences to mitigate noise and partial overlaps, proposing hybrid features integration (global position, local neighborhood, and local surface information) following an iterative scheme. OMNet~\cite{Xu2021} combines global and local learned features from both point clouds in an iterative process to improve partial view registration. Furthermore, more recent studies have explored the use of transformers in PCR, such as GeoTransform \cite{Qin2023}, EGST \cite{Yuan2024}, and REGTR~\cite{Yew2022}, maintaining the focus on learning precise correspondences for subsequent SVD application and also within a constrained transformation range.

Our approach adheres to the current end-to-end trend for PCR, but eliminates the requirement for extensive labeling of correspondences between point clouds for model training. This is feasible because our method can learn correspondences in an unsupervised way, only needing information about the alignment of the point clouds. Moreover, our network combines the processes of identifying matches and their confidence scores into a single step through the novel ``Softmax-Pooling'' layer. The preliminary global matches are subsequently refined in a final stage using a target-guided denoising and registration process. This refinement is carried out considering the features learned by the network of the $K$ nearest neighbors of each of the points to be registered. These are then projected into Euclidean space and employed to solve a weighted least squares problem, thereby estimating the final transformation.

Finally, it is important to highlight that current methods focus on outperforming the state of the art on standard datasets, considering only the limited range of transformations they define. This practice makes the researchers focus on PCR with small transformations ($<$45$^\circ$), where current methods excel~\cite{Wang2019a,choy2020,Lu2021,Cao2021}. The challenge of PCR under significant transformations, despite being a frequent occurrence in real-world data, remains largely unexplored. Furthermore, assessing performance in these more demanding conditions provides insights into the network's ability to generalize across diverse transformations and its rotation invariance, or whether it tends towards overfitting to the training/validation sets.

\section{Proposed approach} 
\label{sec:proposal}

\subsection{Problem formalization}
\label{sec:formalization}

The problem of registration involves two point clouds: the source set $\mathit{X} = \{x_1, x_2, ..., x_m\} \in \mathbb{R}^3$ and the target set $\mathit{Y} = \{y_1, y_2, ..., y_n\} \in \mathbb{R}^3$. The objective is to discover a spatial transformation $\mathit{T} \in \mathcal{T}$ that optimally aligns $\mathit{X}$ to $\mathit{Y}$. This is achieved by minimizing the distance between these two sets, effectively ensuring that the source set is registered or matched as closely as possible to the target set. This can be mathematically expressed as follows: 

\begin{equation} \label{eq:registration}
     \mathbf{T^*} = \underset{\mathit{T} \in \mathcal{T}}{\mathrm{argmin}}~\left\|\mathit{T}(\mathit{X})- \mathit{Y}\right\|^{2}
\end{equation}


Current methods~\cite{choy2020,Lu2021,Cao2021,Wang2019b} address this problem by learning a function that maps each point in $\mathit{X}$ to its corresponding point in $\mathit{Y}$. This function is usually a layer $l$, parameterized by weights $\xi$, that outputs a mapping matrix along with a confidence score for each match, denoted as $l(\mathit{X}, \mathit{Y} | \xi)$. Therefore, this strategy retains all points of $\mathit{X}$, even in the absence of a perfect match, and depends on the confidence score to reduce the contribution of incorrect matches.
The primary rationale for learning confidence scores instead of directly selecting the pairs with the highest scores---i.e.,  $\mathit{x}_\mathit{k}, \mathit{y}_\mathit{k}= \mathrm{arg max}_k l(\mathit{X}, \mathit{Y} | \xi)$---is that the latter is not differentiable. Additionally, this approach necessitates training with ground-truth correspondences, which can be challenging to acquire with real data, especially considering that two real scans might yield different point locations for the similar real-world coordinates.

Our proposal seeks to integrate the steps of matching and confidence estimation, utilizing confidence scores to identify good matches. This enables the network to retain the best pairings, rather than simply mapping one point set to the other, and also facilitates end-to-end training, unlike previous methods that require separate networks for feature matching and confidence estimation. Furthermore, it allows for the \textbf{unsupervised learning} of the confidence scores, guided solely by the alignment error of the point clouds. For this, as proposed in~\cite{choy2020,Cao2021}, we similarly reformulate Equation~\ref{eq:registration} in order to obtain the rotation $\mathbf{R^*}$ and translation $\mathbf{t^*}$ as follows: 

\begin{equation} \label{eq:registration_least_squares}
    \mathbf{R^*},\mathbf{t^*} \, = \, 
    \underset{\mathit{R},\mathit{t}}{\mathrm{argmin}}    
    \left( 
    \sum_{\mathit{i}}^{\mathit{M}} 
    \sum_{\mathit{j}}^{\mathit{N}} 
    \mathit{w}_\mathit{i,j} 
    \left\|\mathit{Rx}_{i}+\mathit{t}-\mathit{y_{j}}\right\|^{2} 
    \right)
\end{equation}

\noindent
where $\mathit{x_i}$ and $\mathit{y_j}$ are the points from the source and target sets, $\mathit{w_i,j}$ is the confidence score, and $\left\| \cdot \right\|$ denotes $\mathit{L_2}$ norm. 

Equation~\ref{eq:registration_least_squares} defines the classic orthogonal Procrustes problem which can be solved in a closed form using weighted SVD. However, to solve this equation, the correspondences between the points in $\mathit{X}$ and $\mathit{Y}$ must be found, such that $\mathit{X}',\mathit{Y}'= \mathrm{argmin}_{\mathit{X},\mathit{Y}} \left \| R\mathit{X}+\mathit{t}-\mathit{Y} \right \|$ is minimized, where $\mathit{X}'$ and $\mathit{Y}'$ are the corresponding points, $\mathit{X}' \subset \mathit{X}$, $\mathit{Y}' \subset \mathit{Y}$, $|\mathit{X}'| = |\mathit{Y}'|$, and $|\cdot|$ is the cardinality of a set.

\begin{figure}[!ht]
    \centering
    \includegraphics[width=\textwidth]{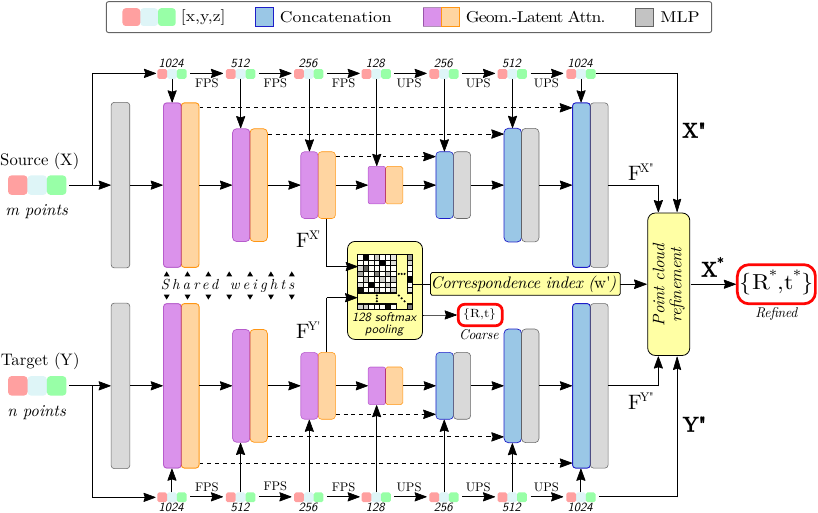}
    \caption{Proposed architecture for 3D rigid PCR. The two optimized loss functions are marked in red. The proposed novel parts of the architecture are highlighted in yellow. FPS and UPS refer to Farthest Point Sampling and Up-Sampling, respectively. Details are described in Section~\ref{sec:network}.}
    \label{fig:architecture}
\end{figure}

To solve Equation~\ref{eq:registration_least_squares} and find good correspondences, we introduce the ReLaTo network architecture, as illustrated in Figure~\ref{fig:architecture}. This is composed of two siamese networks that process the data to extract robust and rotation-invariant descriptors of the points. These descriptors (also called features or embeddings) are first used to find correspondences between pairs of points and sample the most reliable pairs to compute a global and coarse match. Given that both source and target point sets may contain noise, the initially matched points may not align perfectly after the coarse adjustment. Therefore, in a subsequent stage, these points are refined by ``denoising'' them using local information, giving as a result a better registration. Detailed explanations of each of these steps are provided in the following sections.

\subsection{Network architecture}
\label{sec:network}

Our proposal consists of two siamese encoder-decoder networks with residual connections (see Figure~\ref{fig:architecture}). Each network receives as input a point cloud (with the coordinates of the source $\mathit{X}$ and target $\mathit{Y}$ sets, but could easily be extended to include color information) and outputs the set of encoded features ($\mathit{F}^{\mathit{X}'}$ and $\mathit{F}^{\mathit{Y}'}$). Each encoder block is composed of two ResNet layers, which are based on the Geometric-Latent Attention (Ge-Latto) layer proposed in~\cite{Cuevas-Velasquez2021}. We chose this layer because it is light and has been shown to better capture local relationships inside a radius $\mathit{r}$---which represents its receptive field---using geometric and latent information. Each encoder block also sub-samples the point clouds using Farthest Point Sampling (FPS) while increasing the Ge-Latto radius after each block. This ensures an approximately uniform point distribution and obtains local-to-global information~\cite{Cuevas-Velasquez2021}. Although this technique is time-consuming, as it needs to compare all points to find the farthest one, it is possible to store the distance matrix for all points. Therefore, accessing them via index would not have an impact on the execution time of the method. We also follow \cite{Cuevas-Velasquez2021} for the number of sampling points per layer, adopting a sequence of $1024 \rightarrow 512 \rightarrow 256$, and corresponding radius values of $0.1$, $0.2$, and $0.4$. 

According to \cite{Cuevas-Velasquez2021}, this configuration is most effective for points within the range of $[-2, 2]$ in each axis. Thus, we resized the coordinates of each point cloud to this range for training, and scaled up to the original size when computing the error for the evaluation. 

As a result of this sampling, the last layer of the encoder obtains 256 points for $\mathit{X}$ and $\mathit{Y}$, each point with features $\mathit{F}^\mathit{X'_i} \in \mathbb{R}^{D}$ and $\mathit{F}^\mathit{Y'_i} \in \mathbb{R}^{D}$. The features corresponding to these 256 selected points are then processed through a novel layer, termed Softmax Pooling (central yellow box in Figure~\ref{fig:architecture}, better described in Section~\ref{sec:maxPool}) to obtain the confidence scores $w_k$ and a coarse registration with the best $\mathit{K}$ pairs $\{(\mathit{x_1},\mathit{y_1}), (\mathit{x_2},\mathit{y_2}),...,(\mathit{x_k},\mathit{y_k})\}$ (with $\mathit{K}=128$ in the experiments).

The network's intermediate layer consists of one more Ge-Latto layer that processes the 128 points after the last FPS. A mirror architecture is then added for the decoder stage that Up-Samples (UPS) the points until recovering the same initial 1024 points. Tri-linear interpolation and a Multilayer Perceptron (MLP) are used to up-sample the latent features, similar to~\cite{lin2020,Cuevas-Velasquez2021}. This feature up-sampling process refines the coarse registration from the Softmax Pooling layer. 

The following sections provide a detailed explanation of the pooling layer, the procedures performed for point cloud refinement, and the loss function used by the network.

\subsection{Softmax pooling, bilateral consensus and coarse registration}
\label{sec:maxPool}

The proposed Softmax pooling layer is responsible for sampling pairs with high similarities and generating their confidence scores. As stated before, the last layer of the encoder returns 256 sub-sampled points $\mathit{X}' \subset \mathit{X}$ and $\mathit{Y}' \subset \mathit{Y}$, and their respective latent features $F^{\mathit{X}'}$ and $F^{\mathit{Y}'}$. From them, we compute a similarity matrix $S \in \mathbb{R}^{256 \times 256}$ using the dot product (Eq. \ref{eq:dotproduct}), and select the 128 best matches, which are used later in the refinement step (explained in Section~\ref{sec:refinement}).

 \begin{equation}\label{eq:dotproduct}
 \begin{aligned}
    S_{\mathit{i},\mathit{j}} = (\mathit{F}^{\mathit{X}'}_\mathit{i})^\intercal F^{\mathit{Y}'}_\mathit{j}. 
 \end{aligned}
 \end{equation}
 
Instead of using $\mathit{S}$ to obtain a mapping from one set to the other, we use it to estimate a \textit{bilateral consensus} between both mappings. It is a common practice to normalize the similarity matrix either by its columns or rows using a Softmax function, and to use it as a soft-mapping matrix to map $\mathit{X} \mapsto \mathit{Y}$ or $\mathit{Y} \mapsto \mathit{X}$, depending on the normalized dimension. This normalization allows weighting the contribution of each point depending on its score. The main issue with this directional operation is that it cannot guarantee reverse similarity matching. However, correct and robust correspondences must have a bilateral consensus. In our proposal, both directional operations must have a high score for a given pair of points to be considered a correct match. Equation~\ref{eq:bilateral_consensus} illustrates the computation of the bilateral consensus matrix $C$ by element-wise multiplying the two Softmax functions $\phi_r(\cdot)$ and $\phi_c(\cdot)$, where each Softmax normalizes the similarity matrix $\mathit{S}$ by their rows and columns, respectively.

\begin{equation}\label{eq:bilateral_consensus}
\begin{aligned}	
    C = \phi_\mathit{r}(\mathit{S})\odot \phi_\mathit{c}(\mathit{S})
\end{aligned}
\end{equation}

Although we found the work of Lu et al.~\cite{Lu2021} also uses bilateral consensus, it has several differences from our proposal. On the one hand, in~\cite{Lu2021}, matches are estimated within a predetermined neighborhood, a constraint that reduces robustness in scenarios involving noisy data, large transformations, or the registration of partial views. In our case, the entire set of data is used in the correspondence estimation. On the other hand, they concatenate the consensus matrix to other previous features and pass them to the next layer. Instead, we use the matrix $\mathit{C}$ to obtain the top $\mathit{K}$ correspondences from $\mathit{X}'$ and $\mathit{Y}'$. 

Considering that the values of the consensus matrix $\mathit{C}$ are in the range $[0, 1]$---where values closer to $1$ indicate a higher likelihood of correct matching---, we could sample the pairs with the highest values. However, as mentioned before, this kind of operation is not differentiable. Therefore, to allow the network to pick better pairs while still being able to train end-to-end, we use the consensus values of the sampled points as our confidence scores $\mathit{w'}$. This $\mathit{w'}$ is replaced in Equation~\ref{eq:registration_least_squares}, which is solved using differentiable weighted SVD (wSVD)~\cite{paszke2017} to find a coarse global transformation (see Figure~\ref{fig:architecture}). 

Note that the proposed loss function (explained in Section~\ref{sec:loss}) focuses solely on minimizing the difference between ground truth rotation $R$ and translation $\mathit{t}$ and the estimation of $\mathit{R_{pred}}$ and $\mathit{t_{pred}}$ using wSVD, while the confidence scores and matchings are obtained in an unsupervised way. This implies that, in the initial stages of training, the network chooses random matches as corresponding pairs, resulting in a large difference between the predicted transformation and the ground truth. As training continues, the network learns that to reduce this difference, the consensus matrix values belonging to good matches must be high. An example can be seen in Figure~\ref{fig:attention_matrices_modelnet40}.

\subsection{Target-guided denoising and registration refinement}
\label{sec:refinement}

Since the encoder uses a sub-sampled set of points, it can only produce a coarse registration. Thus, as a next step, we proceed to refine the registration by ``denoising'' the matched pairs. Considering that the encoder outputs the pair of points $(\mathit{x', y'})$, which is a good but noisy correspondence, we can refine the position of the source point $\mathit{x'}$ by using the information of its neighborhood. 

One potential solution may be to pick the P-nearest neighbors\footnote{We change K-NN to P-NN to avoid confusion with the $K$ used for corresponding pairs.} and apply an average filter to their coordinates, where each point contributes equally. However, this method may introduce more noise because the true position of the point $\mathit{x'}$ may not be at the centroid of the neighborhood. For this reason, we propose to do a weighted average of the neighborhood coordinate information. 

\begin{figure}[!ht]
    \centering
    \includegraphics[width=\textwidth]{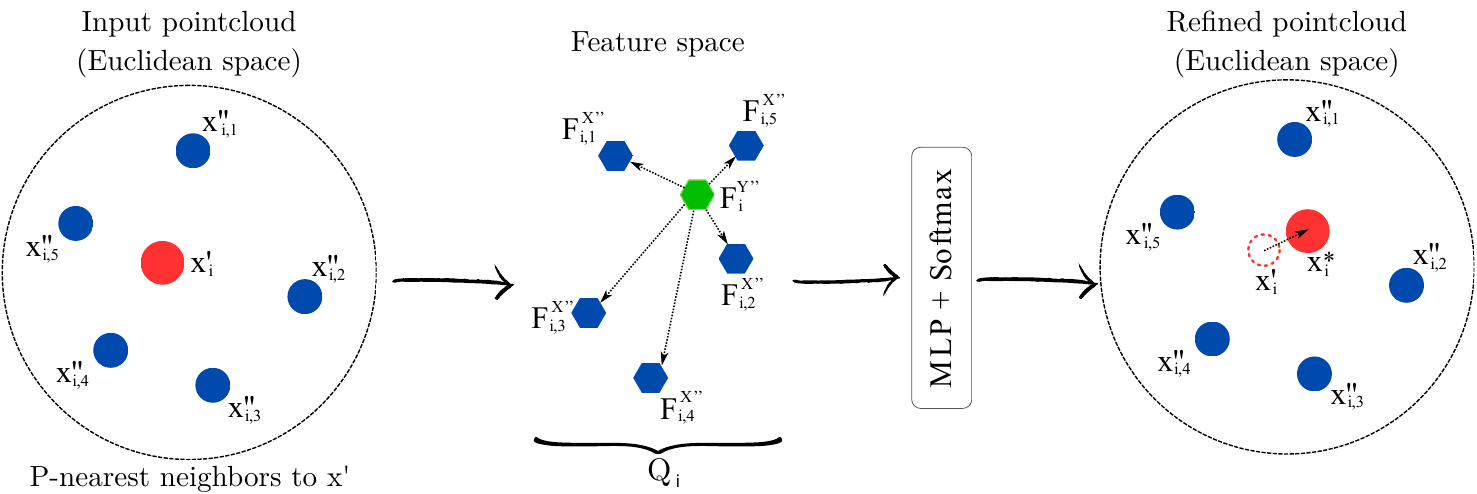}
    \caption{Representation of the target-guided denoising process.}
    \label{fig:refinement}
\end{figure}

The proposed denoising process is composed of three steps (illustrated in Figure~\ref{fig:refinement}). First, for each point $x'$ (from the 128 selected pairs), the P-nearest neighbors ${\mathit{x}''}_\mathit{j}$ (P=15 in the experiments) are found in $\mathit{X}''$ in the Euclidean space, where $\mathit{X}''$ are the up-sampled points obtained by the decoder. Then, the latent features associated with each of the neighbors in $\mathit{X}''$, denoted as $\mathit{F_i}^{\mathit{X}''}$, are grouped in $\mathit{Q_i} \in \mathbb{R}^{P \times \mathit{D}}$, where $\mathit{Q_i} = [\mathit{F}^{\mathit{X}''}_{\mathit{i},1}, \mathit{F}^{\mathit{X}''}_{\mathit{i},2}, \dots, \mathit{F}^{\mathit{X}''}_{\mathit{i},P}]$. Subsequently, as shown in the middle part of Figure \ref{fig:refinement}, the difference in the feature space between the descriptors of $\mathit{Q_i}$ and the descriptor $\mathit{F}^{\mathit{Y}''}_{\mathit{i}}$ of $y'$---the best-matched pair of $x'$---is calculated. Note that $\mathit{F}^{\mathit{Y}''}_{\mathit{i}}$ represents the descriptor of $y'$ obtained by the decoder, which is in the same feature space that $\mathit{Q_i}$. The calculated differences are passed through an MLP and a Softmax layer to obtain the weight $\theta_\mathit{j}$ of each neighbor point (Equation~\ref{eq:refinement}). For instance, in the central part of Figure \ref{fig:refinement}, it is illustrated that $\mathit{F}^{\mathit{X}''}_{\mathit{i},5}$, $\mathit{F}^{\mathit{X}''}_{\mathit{i},1}$ and $\mathit{F}^{\mathit{X}''}_{\mathit{i},2}$ are closer to $\mathit{F}^{\mathit{Y}'}_{\mathit{i}}$, and, thus, they have stronger influence on the denoising decision ($\theta$ in Eq.~\ref{eq:refinement}). Finally, the 3D point $x'$ is refined by relocating its position $x_i^*$ based on the influence of the P neighbors ${\mathit{x}''}_\mathit{j}$ of ${\mathit{x}}_\mathit{i}$ in the latent space. 

\begin{equation}\label{eq:refinement}
\begin{aligned}
    {\mathit{diff}}_\mathit{i} &= \mathit{Q_i} - \mathcal{M}_P\mathit{F}^{\mathit{Y}''}_\mathit{i} \in \mathbb{R}^{P \times \mathit{D}} 
    \\
    \Theta &= \mathit{Softmax}( \mathit{MLP} ( \mathit{diff}_\mathit{i})) \in \mathbb{R}^{P} 
    \\
    {\mathit{x}^{*}}_\mathit{i} &= \sum^{\mathit{P}}_{\mathit{j}=1} \theta_\mathit{j} {\mathit{x}''}_\mathit{j} \\
\end{aligned}
\end{equation}

\noindent
where the operator $\mathcal{M}_P$ replicates the vector P times. The refined points $\mathit{X}^*$ and the set $\mathit{Y}''$ are then used to compute the refined registration using wSVD.

\subsection{Loss}
\label{sec:loss}

The proposed loss function minimizes the difference between the ground truth rotation $R_{gt}$ and translation $t_{gt}$ with the predicted rotation $R_{pred}$ and translation $t_{pred}$ as follows: 

\begin{equation} \label{eq:loss}
\begin{aligned}
   \mathcal{L}_{\mathit{rot}} & = \left \| \mathit{R}_{\mathit{gt}}^\intercal \mathit{R_{pred}} - \mathit{I} \right \|_2 \\
   \mathcal{L}_{\mathit{trans}} & = \left \| \mathit{t_{gt}}-\mathit{t_{pred}} \right \|_2 \\
   \mathcal{L}_\mathit{T} & = \mathcal{L}_{\mathit{rot}} + \mathcal{L}_{trans} \\
   \mathcal{L} & = \lambda \mathcal{L}_{\mathit{T_{coarse}}} + (1-\lambda) \mathcal{L}_{\mathit{T_{refine}}} 
\end{aligned}
\end{equation}

\noindent
where $\mathit{I}$ is a $3 \times 3$ identity matrix and $\mathcal{L}_\mathit{T}$ is the combined loss function that considers both rotation and translation errors. To prevent any bias in the objective tasks, equal importance is applied to both $\mathcal{L}_{\mathit{rot}}$ and $\mathcal{L}_{\mathit{trans}}$. 

This loss is used both in the middle part of the network to minimize coarse registration error ($\mathcal{L}_{\mathit{T_{coarse}}}$) and in the last part for the refinement step ($\mathcal{L}_{\mathit{T_{refine}}}$). In the composition of the total objective loss function $\mathcal{L}$, more weight is given to $\mathcal{L}_{\mathit{T_{coarse}}}$, specifically assigning it a weight of $\lambda=0.6$. This is because coarse registration performs the initial matching and alignment process, which is critical to achieving accurate final results. This not only aligns with theoretical expectations but has also been supported by results obtained in preliminary experiments.

\section{Experimental setup} 
\label{sec:experimental_setup}

This section describes the proposed experimental scheme, including the datasets considered, the training parameters, and the metrics used to assess the goodness of the proposal. 

\subsection{Datasets} 

Two datasets commonly used in registration were considered to benchmark the proposal, namely: ModelNet40~\cite{wu2015} and KITTI~\cite{Menze2015}. Each dataset presents its particular challenges, KITTI was selected as a representative example of real datasets, with partial overlapping and noisy data obtained by real sensors, while ModelNet40 contains synthetic models of objects, which represent much cleaner data but are highly symmetrical, making it difficult to find robust descriptors for the registration task. Specific details about the composition of these datasets can be consulted in the original publications.

For training and evaluation with these datasets, we resort to the standard configuration. Specifically, for ModelNet40, the standard train/test split was employed, whereas for KITTI we adhered to the train/val/test division specified by~\cite{Lu2021}. To conduct a fair comparison with state-of-the-art methods and to demonstrate the generalization capabilities of our proposal, the training sets were augmented considering only rotations in the range of $[-45^\circ, 45^\circ]$, aligning with the approach followed by other methods like DGR~\cite{choy2020}, PCAM~\cite{Cao2021}, or HRegNet~\cite{Lu2021}. In the case of ModelNet40, we also add Gaussian noise with $\sigma=0.01$ during the training stage to increase the variability of the models (this was not applied to the test set to ensure objective evaluation results, not influenced by noise).

To compare our approach against the state of the art under large transformation conditions, we randomly rotate the source ground truth considering different angles: $\{45^\circ, 90^\circ, 135^\circ, 180^\circ\}$. We did not modify the translation, as the positions of the source and target are normalized to their center of mass.

Given that each dataset has different characteristics, there are some particularities in the implementation of the source and target generation. For KITTI, rotations were only applied around the $\mathit{Z}$ axis, as illustrated in Figure~\ref{fig:kitti_images}. In the case of ModelNet40, it does not have predefined source and target sets. Therefore, we randomly selected two subsets of points from the original point cloud to serve as the source and target. Then, the source was rotated to the specified angle, with rotations conducted sequentially along the $[\mathit{x,y,z}]$ axes.

\subsection{Training parameters} 

All models were carried out under the same conditions. Training was performed for 400k iterations with a mini-batch size of 32 samples. Adam was used as optimizer with $\beta_1=0.9$, $\beta_2=0.98$, $\epsilon = 1\mathrm{e}{-9}$, and a constant learning rate $\mathit{lr}=1\mathrm{e}{-4}$.

Following the paper~\cite{Cuevas-Velasquez2021}, we randomly sampled 7,168 points from the original point clouds. These were then processed using FPS to select the 1,024 points used for network input. During training, we normalized the coordinates of the point clouds to the range $[-2, 2]$ in each axis, a configuration identified by \cite{Cuevas-Velasquez2021} as optimal. For evaluation, point coordinates were adjusted back to their original scale.

Regarding the network hyperparameters, we set $\mathit{K}=128$ for the number of correspondences selected by the Softmax pooling layer, $P=15$ for the number of nearest neighbors considered by the target-guided denoising process, and establish $\lambda=0.6$ in the loss function to give more weight to the coarse registration as it performs the first alignment and matching, which is critical to achieving accurate final results.

The implementation was built using Python programming language and the public library PyTorch~\cite{paszke2017}. The machine used consists of an Intel(R) Core(TM) i7-8700 CPU @ 3.20 GHz with 16 GB RAM, a NVIDIA GeForce RTX 2070 with 6 GB GDDR6 Graphics Processing Unit (GPU) with the cuDNN library.

\subsection{Metrics} 

For consistency and ease of comparison, the performance metrics applied to both datasets were the translation error (TE) and rotation error (RE), following the definitions proposed by~\cite{choy2020}. TE is calculated as $\mathit{TE} = \left \| \mathit{t_{pred}} - \mathit{t_{gt}}\right \|_2$, and RE is defined as $RE = \text{arccos}\left [ \left ( \mathit{Tr(R_{gt}}^\intercal \mathit{R_{pred}}) -1 \right ) /2 \right ]$. These metrics are appropriate to compare the methods under the same conditions over several transformations, due to the absolute comparison of the values.

\section{Results} 
\label{sec:experiments}

This section presents the experimental results, highlighting the capability of the method on large transformation scenarios while retaining competitive results with the state of the art. First, a comparison with the existent methods in the literature is presented (Sec.~\ref{sec:results:comparison}), followed by an ablation study over several parameters (Sec.~\ref{sec:results:ablation}), such as the performance behavior of the method considering unsupervised point correspondence, symmetric objects, coarse vs. refined prediction, and the impact of different pooling values. The study further includes a comprehensive analysis of confidence scores and an examination of the network's intermediate outputs.

An extended version of the registration results, as well as a brief explanation of the method, can be found at \url{https://youtu.be/4HnzQnwF00k}.

\subsection{Comparison with existing methods}
\label{sec:results:comparison}

In this section, we compare the proposed method with the current state-of-the-art and classical methods. All networks were tested using their own pre-processing pipeline and the configuration proposed in each original article. We also used the pre-trained weights provided by the authors. Therefore, as explained before, the only variation was the rotation of the source point clouds of the test set, which varied between $45^\circ$ and $180^\circ$ with an increment of $45^\circ$. Note that all methods (including our proposal) were trained under the same conditions, with a limited set of transformations, with rotations in the range $[-45^\circ, 45^\circ]$. This evaluation is designed to assess the generalization capabilities of the methods under large transformations and to determine whether the learned descriptors exhibit rotation invariance.

\begin{figure}[!ht]
    \centering
    \includegraphics[width=0.99\textwidth]{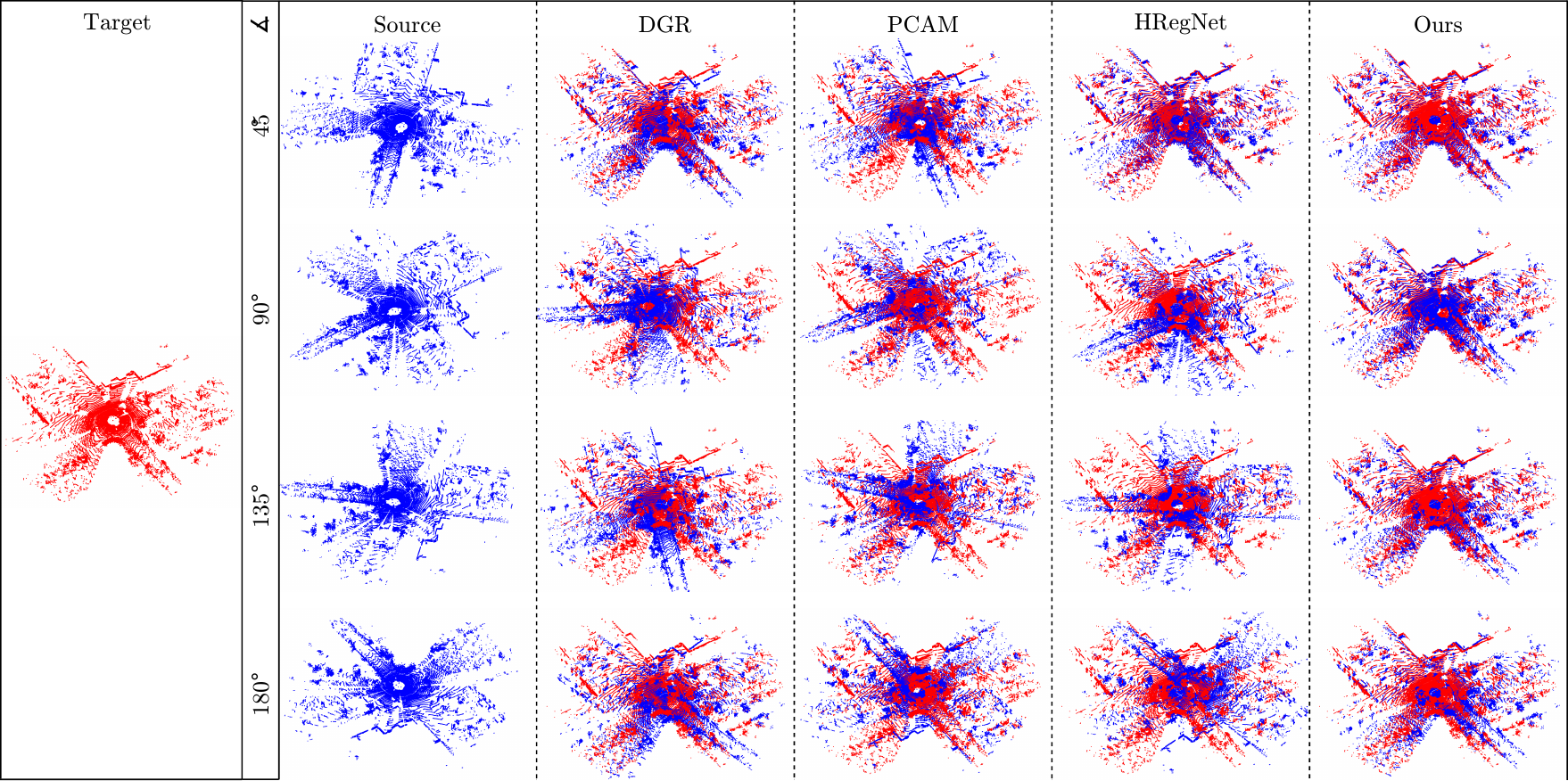}
    \caption{KITTI dataset registrations. The proposed method successfully registers the point sets, even under large transformations, where other current methods fail most of the time.}
    \label{fig:kitti_images}
\end{figure}

\noindent\textbf{KITTI:} 
As observed in Figure~\ref{fig:kitti_rotation_errors}, the DGR~\cite{choy2020}, PCAM~\cite{Cao2021}, and HRegNet~\cite{Lu2021} networks perform well when the rotation varies between $[-45^\circ, 45^\circ]$. However, their performance decreases noticeably when the rotation exceeds this range. In contrast, the error of the proposed method remains consistent across the entire range of transformations. This observation is supported by the data presented in Table~\ref{tab:comp_kitti}, where the ``Avg.'' column shows both the average outcome and the standard deviation of the errors, highlighting the superior rotational invariance of our network compared to existing state-of-the-art solutions. A qualitative comparison is provided in Figure~\ref{fig:kitti_images}, illustrating these distinctions.

\begin{table}[!ht]
    \caption{KITTI results. The table shows the rotation and translation errors under different rotations. It also shows the mean and std. of the rotations. Our network outperforms all the current methods and has less variation in the errors at the different rotation levels considered.}
    \label{tab:comp_kitti}
\centering
\renewcommand{\arraystretch}{1.3}
\setlength{\tabcolsep}{1pt}
\begin{adjustbox}{width = \textwidth, keepaspectratio}
\begin{tabular}{@{\extracolsep{4pt}}crrrrrrrrrrrrrrrrrr} 
\hline
\multirow{3}{*}{Methods} & \multicolumn{9}{c}{RE (deg)}                                                                                                                                                                                                                                                                              & \multicolumn{9}{c}{TE (m)}                                                                                                                                                                                                                                                                                 \\ 
\cline{2-10} \cline{11-19}
                         & \multicolumn{8}{c}{Src-target rotation level}                                                                                                                                                                      & \multicolumn{1}{c}{\multirow{2}{*}{Avg.}}  & \multicolumn{8}{c}{Src-target rotation level}                                                                                                                                                                      & \multicolumn{1}{c}{\multirow{2}{*}{Avg.}}   \\ 
\cline{2-9} \cline{11-18}
                         & \multicolumn{1}{c}{-180} & \multicolumn{1}{c}{-135} & \multicolumn{1}{c}{-90} & \multicolumn{1}{l}{-45} & \multicolumn{1}{l}{45} & \multicolumn{1}{l}{90} & \multicolumn{1}{l}{135} & \multicolumn{1}{l}{180} & \multicolumn{1}{c}{}                          & \multicolumn{1}{l}{-180} & \multicolumn{1}{l}{-135} & \multicolumn{1}{l}{-90} & \multicolumn{1}{l}{-45} & \multicolumn{1}{l}{45} & \multicolumn{1}{l}{90} & \multicolumn{1}{l}{135} & \multicolumn{1}{l}{180} & \multicolumn{1}{c}{}                                            \\ 
\cline{1-1}\cline{2-9} \cline{10-10} \cline{11-18} \cline{19-19}
ICP                  & 173.1                   & 151.6                   & 90.1                   & 22.2                   & 24.8                  & 87.5                    & 145.0                  & 173.1                  & 109.0$\pm$58.3                & 11.7                                      & 12.0                    & 12.5                    & 9.0                    & 9.8                     & 12.5                   & 11.9                   & 11.7                    & 11.4$\pm$1.2                                                                                 \\
HRegNet                  & 174.6                   & 148.7                   & 99.2                   & 10.3                   & 15.7                  & 97.8                  & 153.1                  & 174.6                  & 109.2$\pm$62.0                                          & 10.1                    & 9.2                     & 8.6                    & 2.6                    & 3.5                   & 7.9                   & 8.6                    & 10.0                   & 7.6$\pm$2.7                                           \\
DGR                      & 131.9                   & 127.1                   & 74.4                   & 2.8                    & 3.4                   & 73.9                  & 128.5                  & 131.9                  & 84.2$\pm$52.1                                          & 8.0                     & 12.1                    & 12.4                   & \textbf{0.7}           & \textbf{1.0}          & 12.5                  & 10.9                    & 8.0                    & 8.2$\pm$4.6                                           \\
PCAM                     & 173.8                   & 151.7                   & 82.2                   & 3.6                    & 5.0                   & 82.6                  & 155.1                  & 173.6                  & 103.5$\pm$66.5                                          & 12.5                    & 10.9                    & 10.9                   & 3.7                    & 5.1                   & 12.0                  & 11.6                   & 12.5                   & 9.9$\pm$3.2                                           \\ 
\hline
ReLaTo (Ours)            & \textbf{3.2}            & \textbf{2.6}            & \textbf{1.2}           & \textbf{0.9}           & \textbf{0.8}          & \textbf{1.1}          & \textbf{2.3}           & \textbf{3.7}           & \textbf{2.0$\pm$1.1}                                 & \textbf{1.9}            & \textbf{1.9}            & \textbf{1.6}           & 1.4                    & 1.4                   & \textbf{1.5}          & \textbf{1.6}           & \textbf{1.9}           & \textbf{1.5$\pm$0.2}                                  \\
\hline
\end{tabular}
\end{adjustbox}
\end{table}

\begin{figure}[!ht]
  \centering
  \begin{tabular}{c}
     \includegraphics[width=.8\textwidth]{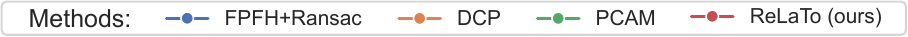} \\ 
     \vspace{0.2cm}
     \includegraphics[width=.95\textwidth]{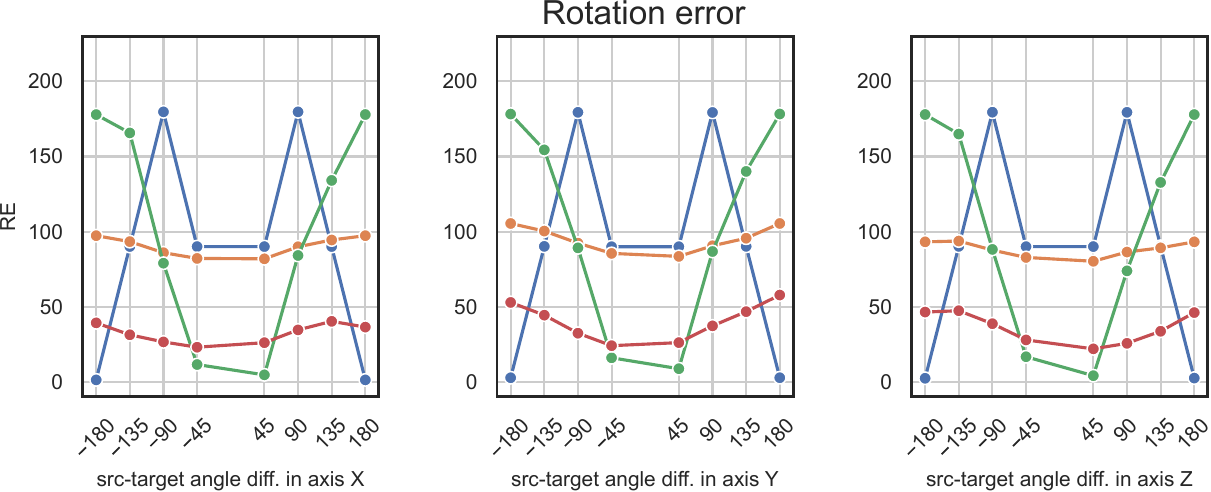} \\
     \vspace{0.2cm}
     \includegraphics[width=.95\textwidth]{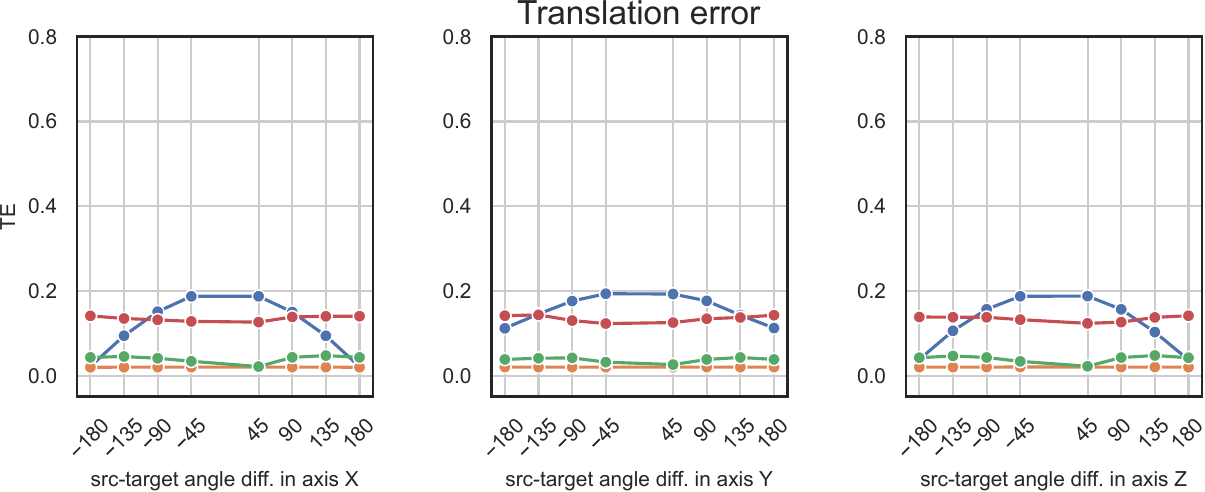}
  \end{tabular}
  \label{fig:kitti_comp}
  \caption{Performance on ModelNet40. Comparison between the proposed model and the state-of-the-art methods under different rotations in the $\mathit{X}$, $\mathit{Y}$, and $\mathit{Z}$ axes. The first row shows the median rotation error (RE) and the second the median translation error (TE).}
  \label{fig:modelnet40_rotation_errors}
\end{figure}

\noindent\textbf{ModelNet40:} 
As shown in Figure~\ref{fig:modelnet40_rotation_errors}, the rotation error of our proposed network shows minimal variation with increasing rotation angles, similar to previous observations with the KITTI dataset. This contrasts with the performance of state-of-the-art methods, such as DCP~\cite{Wang2019a} and PCAM~\cite{Cao2021}, which exhibit larger error fluctuations with increasing degrees of rotation. As seen in Table~\ref{tab:comp_modelnet}, we outperform the state of the art with the lowest rotation error. One downside is that the translation performs slightly worse. However, this error is small and cannot be perceived by visual inspection, as shown in Figure~\ref{fig:attention_matrices_modelnet40}. Note that the ModelNet40 dataset contains some symmetrical objects, such as bottles, bowls, cups, and tables, which can cause large errors even though the registration result visually looks correct (this is discussed in more detail in the ablation study on symmetric objects conducted in Section~\ref{sec:modelnet40_symmetry}). Also note that, to mitigate the impact of symmetry, we report median values for both rotation error (RE) and translation error (TE), which explains why the error figures are higher compared to those in the KITTI dataset.

\begin{table}[!ht]
    \caption{ModelNet40 MAE comparison. The average RE and TE per axis across all rotation levels is shown, as well as their standard deviation. For each rotation level, the RE and TE are calculated using the median error. The table shows that the proposed network has the lowest rotation error with a small standard deviation. Because we use a small number of points for the registration, our translation error is bigger. However, this error is small and cannot be seen by visual inspection (see Figure \ref{fig:attention_matrices_modelnet40}).}
    \label{tab:comp_modelnet}
\centering
\renewcommand{\arraystretch}{1.3}
\resizebox{\textwidth}{!}{
\begin{tabular}{@{\extracolsep{4pt}}crrrrrr} 
\hline
\multirow{2}{*}{\begin{tabular}[c]{@{}c@{}}\\Methods\end{tabular}} & \multicolumn{3}{c}{Average RE per axis (deg)}                                        & \multicolumn{3}{c}{Average TE per axis (m)}                                           \\ 
\cline{2-4}\cline{5-7}
                                                                   & \multicolumn{1}{c}{Axis x} & \multicolumn{1}{c}{Axis y} & \multicolumn{1}{c}{Axis z} & \multicolumn{1}{c}{Axis x} & \multicolumn{1}{c}{Axis y} & \multicolumn{1}{c}{Axis z}  \\ 
\cline{1-1} \cline{2-2}\cline{3-3}\cline{4-4}\cline{5-5}\cline{6-6}\cline{7-7} 
FPFH+Ransac                                                        & 90.26$\pm$63.10            & 90.54$\pm$62.48            & 90.51$\pm$62.61            & 0.11$\pm$0.06              & 0.16$\pm$0.03              & 0.12$\pm$0.06               \\
DCP                                                                & 90.26$\pm$5.92             & 94.82$\pm$7.87             & 88.28$\pm$4.70             & \textbf{0.02$\pm$0.00}     & \textbf{0.02$\pm$0.00}     & \textbf{0.02$\pm$0.00}      \\
PCAM                                                               & 104.35$\pm$66.25           & 106.45$\pm$63.43           & 104.53$\pm$65.44           & 0.04$\pm$0.01              & 0.04$\pm$0.01              & 0.04$\pm$0.01               \\ 
\hline
ReLaTo (Ours)                                                      & \textbf{32.15$\pm$6.05}    & \textbf{40.16$\pm$11.51}   & \textbf{35.93$\pm$9.51}    & 0.14$\pm$0.01              & 0.13$\pm$0.01              & 0.13$\pm$0.01               \\
\hline
\end{tabular}
}
\end{table}

\subsection{Ablation study}
\label{sec:results:ablation}

Following the general evaluation of ReLaTo, this section offers a thorough analysis of the results, exploring the network's performance across diverse conditions to derive insights and conclusions from the observed behaviors.

\subsubsection{Unsupervised point correspondence}

Recent works like DGR, HRegNet, and PCAM adopt a two-step training approach. Initially, they have a backbone network to learn the weights that minimize the distance between the features of the source and target points. Then, they freeze this backbone and learn the transformation between both point sets and the probability of two points being a true match (i.e., the confidence matrix). However, for real data, it is difficult to have ground truth correspondence due to the noisy nature of the data, hence some methods label as true correspondences the points whose distance is under a given threshold. 

In our proposal, the confidence scores are learned in an unsupervised manner by the decoder, which computes a confidence matrix between the source and target point clouds. This matrix is used in two ways. First, it is used by the Softmax-pooling layer to choose the 128 pairs of points with the highest scores. Second, the scores of the 128 points are used as weights to solve the weighted Procrustes problem. This gives the proposed network two advantages compared with previous ones: 1) It does not have to be trained in stages, and 2) it does not need ground truth correspondence (confidence) scores as they are learned as a consequence of the steps described in Section~\ref{sec:maxPool}.

\begin{figure}[!ht]
    \centering
    \includegraphics[width=\textwidth]{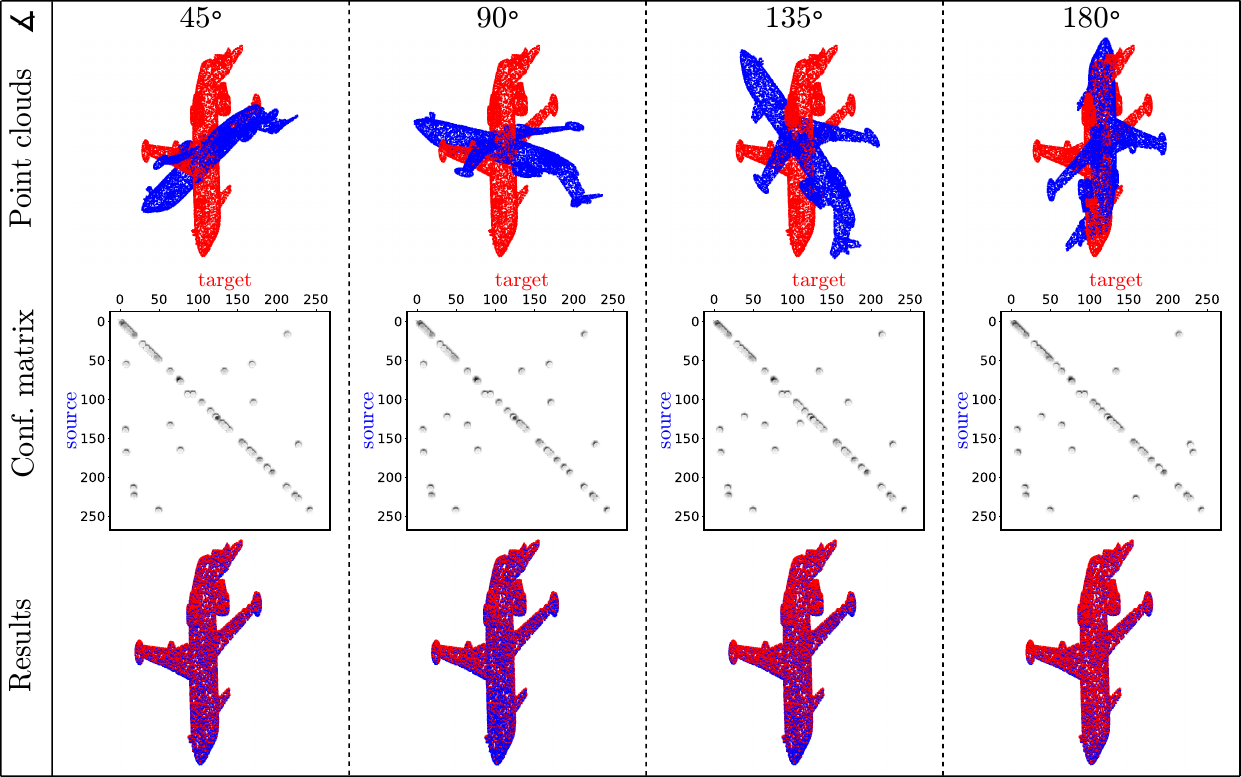}
    \caption{Confidence matrix evaluation. The first row shows the \textcolor{blue}{source}, rotated at different angles, and the \textcolor{red}{target}. The second row shows the confidence probabilities between the source and target points. The darker the point, the higher the probability; only values $>0.08$ are shown for better visualization. The last row includes the registered point clouds.}
    \label{fig:attention_matrices_modelnet40}
\end{figure}

An analysis of the confidence scores learned by our proposal for ModelNet40 is shown in the middle part of Figure~\ref{fig:attention_matrices_modelnet40}. To obtain the true confidence scores we randomly chose a subset of points from the original point cloud as the target, and rotated it to be the source. If the network predicts correctly the correspondence between both sets, then the diagonal values of the learned confidence matrix will be close to 1. As seen, the network is capable of finding correct correspondences between the source and target points even when there are large rotations. The diagonal of the confidence matrix has consistently most of the weight values across all rotations, demonstrating the tolerance of the features to large transformations. We also highlight that during the training process, we randomly sampled two subsets with 7,168 points from the original point cloud, one for the source and the other for the target, as well as adding Gaussian noise. 
Thus, the network becomes accustomed to seeing imperfect correspondences during training.

\subsubsection{Coarse and refined performance}

In this section, the result of each of the network outputs is studied separately. For this, the registration obtained by the pairs chosen by the Softmax pooling layer (coarse registration) has been compared with the output after denoising the point cloud (refined registration). In this case, we will focus on the KITTI dataset, as it presents a greater challenge by containing partial views and a higher level of noise. 

Table~\ref{tab:ablation_loss} shows that the refinement step reduces the translation error by between 0.2 and 0.1 meters on every rotation level. Furthermore, the closeness of coarse and fine rotation errors suggests effective initial match identification by the Softmax pooling layer, as seen for the ModelNet40 dataset in Figure~\ref{fig:attention_matrices_modelnet40}.  This similarity indicates that the initial matches are of high quality and that the collaborative effect of both the coarse and refinement losses significantly enhances the final outcome.

\begin{table}[!ht]
    \caption{Results in terms of RE and TE comparing the coarse and refined registration for the KITTI dataset.}
    \label{tab:ablation_loss}
\small
\centering
\renewcommand{\arraystretch}{1.3}
\resizebox{1\textwidth}{!}{
\begin{tabular}{@{\extracolsep{4pt}}ccccccccccccccccccc} 
\hline
\multirow{3}{*}{Loss}   & \multicolumn{9}{c}{RE (deg)}          & \multicolumn{9}{c}{TE (m)}                                                                                                                                                                                                                                         \\ 
\cline{2-10}\cline{11-19}
                          & \multicolumn{8}{c}{Rotations src-target}   & \multicolumn{1}{c}{\multirow{2}{*}{Avg.}} 
                          & \multicolumn{8}{c}{Rotations src-target}   & \multicolumn{1}{c}{\multirow{2}{*}{Avg.}}   \\ 
\cline{2-9}\cline{11-18}
                          & -180 & -135 & -90  & \multicolumn{1}{l}{-45} & \multicolumn{1}{l}{45} & \multicolumn{1}{l}{90} & \multicolumn{1}{l}{135} & \multicolumn{1}{l}{180} &                           & \multicolumn{1}{l}{-180} & \multicolumn{1}{l}{-135} & \multicolumn{1}{l}{-90} & \multicolumn{1}{l}{-45} & \multicolumn{1}{l}{45} & \multicolumn{1}{l}{90} & \multicolumn{1}{l}{135} & \multicolumn{1}{l}{180}                  &                            \\ 
\cline{1-1} \cline{2-9}\cline{10-10}\cline{11-18} \cline{19-19}
Coarse                    & 3.3  & 2.6  & 1.2  & 0.9                     & 0.8                    & 1.1                    & 2.3                     & 3.7                     & 2.0$\pm$1.1                       & 2.1                      & 2.1                      & 1.8                     & 1.6                     & 1.5                    & 1.6                    & 1.7                     & 2.1                     & 1.8$\pm$0.2                        \\
Refined                   & 3.3  & 2.6  & 1.2  & 0.9                     & 0.8                    & 1.2                    & 2.3                     & 3.8                     & 2.0$\pm$1.1                       & 1.9                      & 1.9                      & 1.6                     & 1.4                     & 1.4                    & 1.4                    & 1.6                     & 2.0                     & 1.7$\pm$0.2                        \\ 
\hline
\end{tabular}
}
\end{table}

\subsubsection{Different pooling values}

In this section, we delve into another important aspect of the architecture: the pooling size, denoted by the parameter $K$, used by the Softmax pooling layer.

Table~\ref{tab:ablation_softmax} shows that selecting $\mathit{K}$=32 pairs of points from the confidence matrix is good enough to have registration with low errors. The best performance can be found between $\mathit{K}$=64 and $\mathit{K}$=128, where there is a good trade-off between the rotation and translation errors. Conversely, settings of $\mathit{K}=256$ and beyond result in increased errors, indicating that higher values of $\mathit{K}$ do not necessarily enhance registration accuracy.

No significant improvement is found with the modification of $\mathit{K}$ over the rotation error. On the other hand, an increase in the value of $\mathit{K}$ implies incorporating more information, leading to worst absolute results in translation, with a consistent decline observed, reaching an average absolute difference of 0.4 meters when comparing $\mathit{K}=32$ to $\mathit{K}=256$. In this scenario, the choice of a lower $\mathit{K}$ is more suitable. Intuitively, the more pairs are added, the more outliers and incorrect matches may be included.

\begin{table}[!ht]
    \caption{Results comparing the performance of the network in terms of RE and TE when varying the pool size ($K$ parameter) of the Softmax pooling layer for the KITTI dataset.}
    \label{tab:ablation_softmax}
\small
\centering
\renewcommand{\arraystretch}{1.3}
\resizebox{1\textwidth}{!}{
\begin{tabular}{@{\extracolsep{4pt}}ccccccccccccccccccc} 
\hline
\multirow{3}{*}{$K$ value}   & \multicolumn{9}{c}{RE (deg)}          & \multicolumn{9}{c}{TE (m)}     \\ 
\cline{2-10}\cline{11-19}
                          & \multicolumn{8}{c}{Rotations src-target}   & \multicolumn{1}{c}{\multirow{2}{*}{Avg.}} 
                          & \multicolumn{8}{c}{Rotations src-target}   & \multicolumn{1}{c}{\multirow{2}{*}{Avg.}}   \\ 
\cline{2-9}\cline{11-18}
                          & -180 & -135 & -90  & \multicolumn{1}{l}{-45} & \multicolumn{1}{l}{45} & \multicolumn{1}{l}{90} & \multicolumn{1}{l}{135} & \multicolumn{1}{l}{180} &                           & \multicolumn{1}{l}{-180} & \multicolumn{1}{l}{-135} & \multicolumn{1}{l}{-90} & \multicolumn{1}{l}{-45} & \multicolumn{1}{l}{45} & \multicolumn{1}{l}{90} & \multicolumn{1}{l}{135} & \multicolumn{1}{l}{180}                  &                            \\ 
\cline{1-1} \cline{2-9}\cline{10-10}\cline{11-18} \cline{19-19}

\multicolumn{1}{l}{K=32}  & 3.4  & 2.7  & 1.4  & 1.0                     & 1.0                    & 1.3                    & 2.4                     & 3.8                     & 2.1$\pm$1.0                       & 1.8                      & 1.8                      & 1.5                     & 1.3                     & 1.3                    & 1.4                    & 1.6                     & 1.9                     & 1.6$\pm$0.2                        \\
\multicolumn{1}{l}{K=64}  & 3.3  & 2.6  & 1.2  & 0.9                     & 0.8                    & 1.2                    & 2.3                     & 3.8                     & 2.0$\pm$1.1                       & 1.9                      & 1.9                      & 1.6                     & 1.4                     & 1.4                    & 1.4                    & 1.6                     & 2.0                     & 1.7$\pm$0.2                        \\
\multicolumn{1}{l}{K=128} & 3.3  & 2.6  & 1.2  & 0.9                     & 0.8                    & 1.2                    & 2.3                     & 3.8                     & 2.0$\pm$1.1                       & 1.9                      & 1.9                      & 1.6                     & 1.4                     & 1.4                    & 1.4                    & 1.6                     & 2.0                     & 1.7$\pm$0.2                        \\
\multicolumn{1}{l}{K=256} & 3.5  & 2.7  & 1.3  & 0.9                     & 0.9                    & 1.3                    & 2.3                     & 3.6                     & 2.1$\pm$1.1                       & 2.3                      & 2.3                      & 2.0                     & 1.9                     & 1.7                    & 1.8                    & 1.9                     & 2.3                     & 2.0$\pm$0.2                        \\
\hline
\end{tabular}
}
\end{table}

\subsection{Symmetric objects in ModelNet40}
\label{sec:modelnet40_symmetry}

\begin{figure}[!ht]
    \centering
    \includegraphics[width=1.\textwidth]{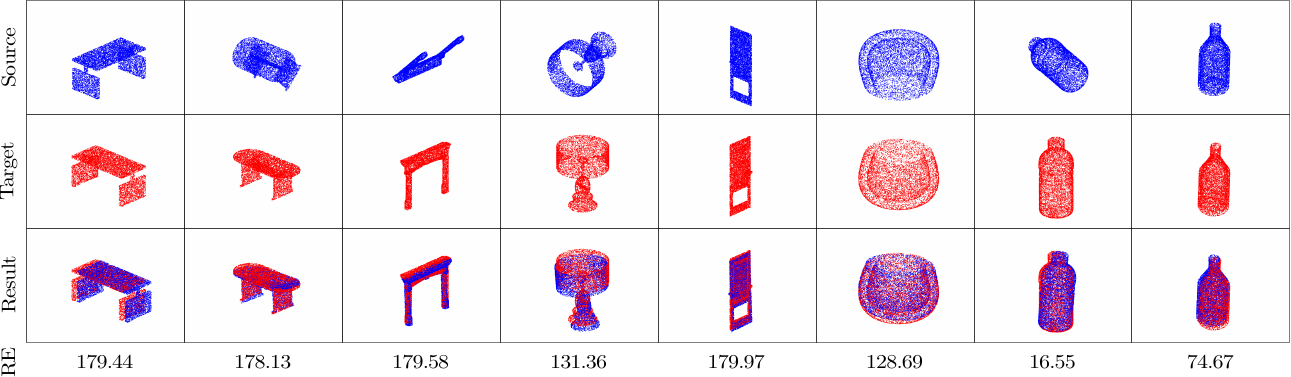}
    \caption{Examples of symmetric objects and their rotation errors (RE).}
    \label{fig:modelnet40_symmetric}
\end{figure}

As previously mentioned in Section~\ref{sec:results:comparison}, the higher error obtained in ModelNet40 is due to the symmetries that some objects in the dataset have. Figure~\ref{fig:modelnet40_symmetric} presents some of these objects and their rotation error (RE). In this figure, we can observe that despite the visually satisfactory appearance of the predicted transformation and point cloud registration, the rotation error is high. 

These objects have the peculiarity of being symmetric, which means that they can be flipped over one or more axis and still remain the same. It can be observed that for some objects that are planar or symmetric, the RE reports a value closer to $180^\circ$. In addition, for some cylindrical objects, the reported error is high even though qualitatively the registration looks good. 

In these cases, the registration method cannot distinguish between the symmetrical positions and, sometimes, returns the mirrored image. In future works, we are considering adding a 'flipped' option network path that considers both possibilities and selects the better fit or including other information, such as color, to reduce this mirroring effect . 

\subsection{Network intermediate outputs}
\label{sec:net_res}

\begin{figure}[!ht]
    \centering
    \includegraphics[width=1\textwidth]{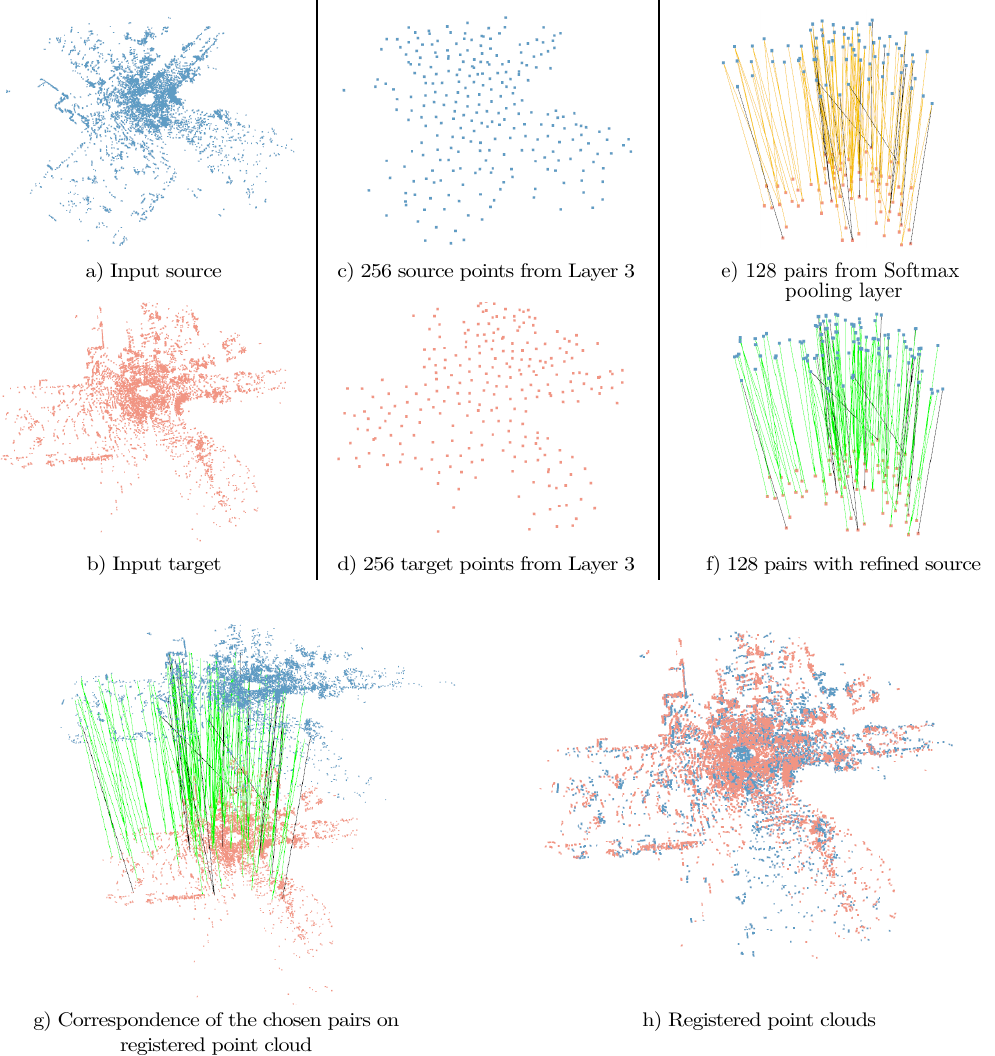}
    \caption{a) and b) show the input point clouds from the KITTI dataset; c) and d) present the 256 sub-sampled points from the last layer of the encoder; e) displays the 128 points picked by the Softmax pooling layer, with lines showing the correspondence between the target and the transformed source (correspondences with low confidence ($<0.05$) are in black); f) shows the output of the last layer of our network, which refines the position of the source points; g) shows the registered input source with the correspondences picked by our network; and h) the registered result.}
    \label{fig:intermediate_outputs}
\end{figure}

Finally, we are going to qualitatively analyze the intermediate outputs provided by the network. Figure~\ref{fig:intermediate_outputs} shows how the network processes the input point clouds as well as the 128 pairs chosen by our Softmax pooling layer (Figure e). As can be seen, these points are homogeneously distributed over the entire surface of the cloud to be registered. The figure illustrates the correspondences between the pairs as colored lines, where the black lines represent the pairs with low confidence scores ($< 0.05$). As can be seen in Figures e), f), and g), the mismatched pairs have low confidence scores, which reduces the contribution of those pairs when solving the weighted Procrustes problem.

\section{Conclusions}
\label{sec:conclusions}

We presented ReLaTo, a learning-based method designed for the precise alignment of two point clouds undergoing large transformations. The proposed method achieves this by learning in an unsupervised manner the confidence scores between the source and target points through bilateral consensus. This confidence score is then passed to a Softmax pooling layer to sample the best pairs, which are used to estimate a coarse registration. The registration is then improved by refining the position of the source points that were matched by the Softmax pooling layer. This is done in a novel target-guided denoising layer that combines the information of the neighbors of each point using a weighted average that is learned through an attention layer.

For experimentation, we analyzed the proposal considering two of the most commonly used datasets for this task. Results show that the confidence scores learned by the proposed model are consistent under large rotation, as well as have a small variation in the error across the different rotation levels. Consistency and low error variability in both rotation and translation regardless of the angle of rotation have been achieved, in contrast to the current state of the art, which only works well for a limited range of transformations, not generalizing well for large transformations. This indicates that the features learned by the network are more tolerant to large transformations than current state-of-the-art methods, even when trained under similar data distribution and rotation ranges. 

For future work, we aim to enrich the point cloud data by incorporating additional attributes such as color. Moreover, we plan to assess the effectiveness of our proposal by exploring alternative backbone architectures for feature extraction, broadening the scope and applicability of our method.

\section*{Acknowledgments}

Marcelo Saval-Calvo was supported by a Valencian Regional Grant (BEST/2021/025). The DGX-A100 cluster at the University of Alicante has been used for the research, funded by the ``Generalitat Valenciana'' and the ``European Union'' through IDIFEDER/2020/003 project.



\bibliographystyle{elsarticle-num}

\bibliography{paper.bib}

\begin{thebibliography}{10}
\expandafter\ifx\csname url\endcsname\relax
  \def\url#1{\texttt{#1}}\fi
\expandafter\ifx\csname urlprefix\endcsname\relax\def\urlprefix{URL }\fi
\expandafter\ifx\csname href\endcsname\relax
  \def\href#1#2{#2} \def\path#1{#1}\fi

\bibitem{Saval-Calvo2015}
M.~Saval-Calvo, J.~Azorin-Lopez, A.~Fuster-Guillo, H.~Mora-Mora, \href{https://linkinghub.elsevier.com/retrieve/pii/S0957417415005503}{{$\mu$-MAR: Multiplane 3D Marker based Registration for depth-sensing cameras}}, Expert Systems with Applications 42~(23) (2015) 9353--9365.
\newblock \href {https://doi.org/10.1016/j.eswa.2015.08.011} {\path{doi:10.1016/j.eswa.2015.08.011}}.
\newline\urlprefix\url{https://linkinghub.elsevier.com/retrieve/pii/S0957417415005503}

\bibitem{Kim2018}
P.~Kim, J.~Chen, Y.~K. Cho, \href{https://linkinghub.elsevier.com/retrieve/pii/S0926580517303990}{{SLAM-driven robotic mapping and registration of 3D point clouds}}, Automation in Construction 89 (2018) 38--48.
\newblock \href {https://doi.org/10.1016/j.autcon.2018.01.009} {\path{doi:10.1016/j.autcon.2018.01.009}}.
\newline\urlprefix\url{https://linkinghub.elsevier.com/retrieve/pii/S0926580517303990}

\bibitem{Chaudhury2020}
A.~Chaudhury, \href{https://linkinghub.elsevier.com/retrieve/pii/S0924271616303987 https://ieeexplore.ieee.org/document/9184264/}{{Multilevel Optimization for Registration of Deformable Point Clouds}}, IEEE Transactions on Image Processing 29 (2020) 8735--8746.
\newblock \href {https://doi.org/10.1109/TIP.2020.3019649} {\path{doi:10.1109/TIP.2020.3019649}}.
\newline\urlprefix\url{https://linkinghub.elsevier.com/retrieve/pii/S0924271616303987 https://ieeexplore.ieee.org/document/9184264/}

\bibitem{Saval-Calvo2018}
M.~Saval-Calvo, J.~Azorin-Lopez, A.~Fuster-Guillo, V.~Villena-Martinez, R.~B. Fisher, \href{http://linkinghub.elsevier.com/retrieve/pii/S1077314218300080 https://linkinghub.elsevier.com/retrieve/pii/S1077314218300080}{{3D non-rigid registration using color: Color Coherent Point Drift}}, Computer Vision and Image Understanding 169 (2018) 119--135.
\newblock \href {https://doi.org/10.1016/j.cviu.2018.01.008} {\path{doi:10.1016/j.cviu.2018.01.008}}.
\newline\urlprefix\url{http://linkinghub.elsevier.com/retrieve/pii/S1077314218300080 https://linkinghub.elsevier.com/retrieve/pii/S1077314218300080}

\bibitem{pomerleau2015}
F.~Pomerleau, F.~Colas, R.~Siegwart, \href{http://www.nowpublishers.com/article/Details/ROB-035}{{A Review of Point Cloud Registration Algorithms for Mobile Robotics}}, Foundations and Trends in Robotics 4~(1) (2015) 1--104.
\newblock \href {https://doi.org/10.1561/2300000035} {\path{doi:10.1561/2300000035}}.
\newline\urlprefix\url{http://www.nowpublishers.com/article/Details/ROB-035}

\bibitem{Villena-Martinez2020}
V.~Villena-Martinez, S.~Oprea, M.~Saval-Calvo, J.~Azorin-Lopez, A.~Fuster-Guillo, R.~B. Fisher, \href{https://www.mdpi.com/2076-3417/10/21/7524}{{When Deep Learning Meets Data Alignment: A Review on Deep Registration Networks (DRNs)}}, Applied Sciences 10~(21) (2020) 7524.
\newblock \href {https://doi.org/10.3390/app10217524} {\path{doi:10.3390/app10217524}}.
\newline\urlprefix\url{https://www.mdpi.com/2076-3417/10/21/7524}

\bibitem{kaljaca2019}
D.~Kaljaca, N.~Mayer, B.~Vroegindeweij, A.~Mencarelli, E.~van Henten, T.~Brox, \href{https://ieeexplore.ieee.org/document/8968446/}{{Automated Boxwood Topiary Trimming with a Robotic Arm and Integrated Stereo Vision}}, in: 2019 IEEE/RSJ International Conference on Intelligent Robots and Systems (IROS), IEEE, 2019, pp. 5542--5549.
\newblock \href {https://doi.org/10.1109/IROS40897.2019.8968446} {\path{doi:10.1109/IROS40897.2019.8968446}}.
\newline\urlprefix\url{https://ieeexplore.ieee.org/document/8968446/}

\bibitem{pu2018}
C.~Pu, N.~Li, R.~Tylecek, B.~Fisher, {DUGMA: Dynamic Uncertainty-Based Gaussian Mixture Alignment}, in: 2018 International Conference on 3D Vision (3DV), IEEE, 2018, pp. 766--774.
\newblock \href {https://doi.org/10.1109/3DV.2018.00092} {\path{doi:10.1109/3DV.2018.00092}}.

\bibitem{Menze2015}
M.~Menze, A.~Geiger, \href{http://ieeexplore.ieee.org/document/7298925/}{{Object scene flow for autonomous vehicles}}, in: 2015 IEEE Conference on Computer Vision and Pattern Recognition (CVPR), IEEE, 2015, pp. 3061--3070.
\newblock \href {https://doi.org/10.1109/CVPR.2015.7298925} {\path{doi:10.1109/CVPR.2015.7298925}}.
\newline\urlprefix\url{http://ieeexplore.ieee.org/document/7298925/}

\bibitem{Zhao2021}
H.~Zhao, Z.~Liang, C.~Wang, M.~Yang, \href{https://ieeexplore.ieee.org/document/9361121/}{{CentroidReg: A Global-to-Local Framework for Partial Point Cloud Registration}}, IEEE Robotics and Automation Letters 6~(2) (2021) 2533--2540.
\newblock \href {https://doi.org/10.1109/LRA.2021.3061369} {\path{doi:10.1109/LRA.2021.3061369}}.
\newline\urlprefix\url{https://ieeexplore.ieee.org/document/9361121/}

\bibitem{Villena-Martinez2021}
V.~Villena-Martinez, M.~Saval-Calvo, J.~Azorin-Lopez, A.~Fuster-Guillo, R.~B. Fisher, \href{https://ieeexplore.ieee.org/document/9533295/}{{Local-Global based Deep Registration Neural Network for Rigid Alignment}}, in: 2021 International Joint Conference on Neural Networks (IJCNN), IEEE, 2021, pp. 1--8.
\newblock \href {https://doi.org/10.1109/IJCNN52387.2021.9533295} {\path{doi:10.1109/IJCNN52387.2021.9533295}}.
\newline\urlprefix\url{https://ieeexplore.ieee.org/document/9533295/}

\bibitem{Yuan2022}
M.~Yuan, X.~Li, L.~Cheng, X.~Li, H.~Tan, \href{https://www.mdpi.com/2079-9292/11/2/263}{{A Coarse-to-Fine Registration Approach for Point Cloud Data with Bipartite Graph Structure}}, Electronics 11~(2) (2022) 263.
\newblock \href {https://doi.org/10.3390/electronics11020263} {\path{doi:10.3390/electronics11020263}}.
\newline\urlprefix\url{https://www.mdpi.com/2079-9292/11/2/263}

\bibitem{Yu2021}
H.~Yu, F.~Li, M.~Saleh, B.~Busam, S.~Ilic, {Cofinet: Reliable coarse-to-fine correspondences for robust pointcloud registration}, in: Advances in Neural Information Processing Systems (NeurIPS), 2021, pp. 23872--23884.

\bibitem{Guo2014}
Y.~Guo, M.~Bennamoun, F.~Sohel, M.~Lu, J.~Wan, \href{https://ieeexplore.ieee.org/document/6787078/}{{3D Object Recognition in Cluttered Scenes with Local Surface Features: A Survey}}, IEEE Transactions on Pattern Analysis and Machine Intelligence 36~(11) (2014) 2270--2287.
\newblock \href {https://doi.org/10.1109/TPAMI.2014.2316828} {\path{doi:10.1109/TPAMI.2014.2316828}}.
\newline\urlprefix\url{https://ieeexplore.ieee.org/document/6787078/}

\bibitem{Yang2016}
J.~Yang, Z.~Cao, Q.~Zhang, \href{https://linkinghub.elsevier.com/retrieve/pii/S0020025516300378}{{A fast and robust local descriptor for 3D point cloud registration}}, Information Sciences 346-347 (2016) 163--179.
\newblock \href {https://doi.org/10.1016/j.ins.2016.01.095} {\path{doi:10.1016/j.ins.2016.01.095}}.
\newline\urlprefix\url{https://linkinghub.elsevier.com/retrieve/pii/S0020025516300378}

\bibitem{Charles2017}
C.~R. Qi, H.~Su, M.~Kaichun, L.~J. Guibas, \href{http://ieeexplore.ieee.org/document/8099499/}{{PointNet: Deep Learning on Point Sets for 3D Classification and Segmentation}}, in: 2017 IEEE Conference on Computer Vision and Pattern Recognition (CVPR), IEEE, 2017, pp. 77--85.
\newblock \href {https://doi.org/10.1109/CVPR.2017.16} {\path{doi:10.1109/CVPR.2017.16}}.
\newline\urlprefix\url{http://ieeexplore.ieee.org/document/8099499/}

\bibitem{Yuan2020}
W.~Yuan, B.~Eckart, K.~Kim, V.~Jampani, D.~Fox, J.~Kautz, \href{https://link.springer.com/10.1007/978-3-030-58558-7_43}{{DeepGMR: Learning Latent Gaussian Mixture Models for Registration}}, in: 2020 European Conference on Computer Vision (ECCV), 2020, pp. 733--750.
\newblock \href {https://doi.org/10.1007/978-3-030-58558-7_43} {\path{doi:10.1007/978-3-030-58558-7_43}}.
\newline\urlprefix\url{https://link.springer.com/10.1007/978-3-030-58558-7_43}

\bibitem{Ao2021}
S.~Ao, Q.~Hu, B.~Yang, A.~Markham, Y.~Guo, \href{https://ieeexplore.ieee.org/document/9577271/}{{SpinNet: Learning a General Surface Descriptor for 3D Point Cloud Registration}}, in: 2021 IEEE/CVF Conference on Computer Vision and Pattern Recognition (CVPR), IEEE, 2021, pp. 11748--11757.
\newblock \href {https://doi.org/10.1109/CVPR46437.2021.01158} {\path{doi:10.1109/CVPR46437.2021.01158}}.
\newline\urlprefix\url{https://ieeexplore.ieee.org/document/9577271/}

\bibitem{Fischler1981}
M.~A. Fischler, R.~C. Bolles, \href{https://dl.acm.org/doi/10.1145/358669.358692}{{Random sample consensus: a paradigm for model fitting with applications to image analysis and automated cartography}}, Communications of the ACM 24~(6) (1981) 381--395.
\newblock \href {https://doi.org/10.1145/358669.358692} {\path{doi:10.1145/358669.358692}}.
\newline\urlprefix\url{https://dl.acm.org/doi/10.1145/358669.358692}

\bibitem{myronenko2010}
A.~Myronenko, {Xubo Song}, \href{http://www.ncbi.nlm.nih.gov/pubmed/20975122 http://ieeexplore.ieee.org/document/5432191/}{{Point Set Registration: Coherent Point Drift}}, IEEE Transactions on Pattern Analysis and Machine Intelligence 32~(12) (2010) 2262--2275.
\newblock \href {https://doi.org/10.1109/TPAMI.2010.46} {\path{doi:10.1109/TPAMI.2010.46}}.
\newline\urlprefix\url{http://www.ncbi.nlm.nih.gov/pubmed/20975122 http://ieeexplore.ieee.org/document/5432191/}

\bibitem{Eggert1997}
D.~Eggert, A.~Lorusso, R.~Fisher, {Estimating 3-D rigid body transformations: a comparison of four major algorithms}, Machine Vision and Applications 9~(5-6) (1997) 272--290.
\newblock \href {https://doi.org/10.1007/s001380050048} {\path{doi:10.1007/s001380050048}}.

\bibitem{choy2020}
C.~Choy, W.~Dong, V.~Koltun, \href{https://ieeexplore.ieee.org/document/9157005/}{{Deep Global Registration}}, in: 2020 IEEE/CVF Conference on Computer Vision and Pattern Recognition (CVPR), IEEE, 2020, pp. 2511--2520.
\newblock \href {https://doi.org/10.1109/CVPR42600.2020.00259} {\path{doi:10.1109/CVPR42600.2020.00259}}.
\newline\urlprefix\url{https://ieeexplore.ieee.org/document/9157005/}

\bibitem{Cao2021}
A.-Q. Cao, G.~Puy, A.~Boulch, R.~Marlet, {PCAM: Product of Cross-Attention Matrices for Rigid Registration of Point Clouds}, in: I2021 EEE/CVF International Conference on Computer Vision (ICCV), 2021, pp. 13229--13238.
\newblock \href {http://arxiv.org/abs/2110.01269} {\path{arXiv:2110.01269}}.

\bibitem{Lu2021}
F.~Lu, G.~Chen, Y.~Liu, L.~Zhang, S.~Qu, S.~Liu, R.~Gu, \href{http://arxiv.org/abs/2107.11992}{{HRegNet: A Hierarchical Network for Large-scale Outdoor LiDAR Point Cloud Registration}}, in: 2021 IEEE/CVF International Conference on Computer Vision (ICCV), 2021, pp. 16014--16023.
\newblock \href {http://arxiv.org/abs/2107.11992} {\path{arXiv:2107.11992}}.
\newline\urlprefix\url{http://arxiv.org/abs/2107.11992}

\bibitem{besl1992}
P.~Besl, N.~D. McKay, \href{http://ieeexplore.ieee.org/document/121791/}{{A method for registration of 3-D shapes}}, IEEE Transactions on Pattern Analysis and Machine Intelligence 14~(2) (1992) 239--256.
\newblock \href {https://doi.org/10.1109/34.121791} {\path{doi:10.1109/34.121791}}.
\newline\urlprefix\url{http://ieeexplore.ieee.org/document/121791/}

\bibitem{Ginzburg2022}
D.~Ginzburg, D.~Raviv, \href{https://ieeexplore.ieee.org/document/9897800/}{{Deep Weighted Consensus Dense Correspondence Confidence Maps for 3d Shape Registration}}, in: 2022 IEEE International Conference on Image Processing (ICIP), IEEE, 2022, pp. 71--75.
\newblock \href {https://doi.org/10.1109/ICIP46576.2022.9897800} {\path{doi:10.1109/ICIP46576.2022.9897800}}.
\newline\urlprefix\url{https://ieeexplore.ieee.org/document/9897800/}

\bibitem{Han2023}
X.-F. Han, Z.-A. Feng, S.-J. Sun, G.-Q. Xiao, \href{https://link.springer.com/10.1007/s10462-023-10486-4}{{3D point cloud descriptors: state-of-the-art}}, Artificial Intelligence Review 56~(10) (2023) 12033--12083.
\newblock \href {https://doi.org/10.1007/s10462-023-10486-4} {\path{doi:10.1007/s10462-023-10486-4}}.
\newline\urlprefix\url{https://link.springer.com/10.1007/s10462-023-10486-4}

\bibitem{Zhang2022}
Y.~Zhang, Q.~Hu, G.~Xu, Y.~Ma, J.~Wan, Y.~Guo, \href{https://ieeexplore.ieee.org/document/9879147/}{{Not All Points Are Equal: Learning Highly Efficient Point-based Detectors for 3D LiDAR Point Clouds}}, in: 2022 IEEE/CVF Conference on Computer Vision and Pattern Recognition (CVPR), IEEE, 2022, pp. 18931--18940.
\newblock \href {https://doi.org/10.1109/CVPR52688.2022.01838} {\path{doi:10.1109/CVPR52688.2022.01838}}.
\newline\urlprefix\url{https://ieeexplore.ieee.org/document/9879147/}

\bibitem{Qian2023}
J.~Qian, D.~Tang, {RRGA-Net: Robust Point Cloud Registration Based on Graph Convolutional Attention}, Sensors 23~(24) (2023) 9651.
\newblock \href {https://doi.org/10.3390/s23249651} {\path{doi:10.3390/s23249651}}.

\bibitem{agamennoni2016}
G.~Agamennoni, S.~Fontana, R.~Y. Siegwart, D.~G. Sorrenti, \href{http://ieeexplore.ieee.org/document/7759602/}{{Point Clouds Registration with Probabilistic Data Association}}, in: 2016 IEEE/RSJ International Conference on Intelligent Robots and Systems (IROS), IEEE, 2016, pp. 4092--4098.
\newblock \href {https://doi.org/10.1109/IROS.2016.7759602} {\path{doi:10.1109/IROS.2016.7759602}}.
\newline\urlprefix\url{http://ieeexplore.ieee.org/document/7759602/}

\bibitem{bouaziz2013}
S.~Bouaziz, A.~Tagliasacchi, M.~Pauly, \href{https://onlinelibrary.wiley.com/doi/10.1111/cgf.12178}{{Sparse Iterative Closest Point}}, Computer Graphics Forum 32~(5) (2013) 113--123.
\newblock \href {https://doi.org/10.1111/cgf.12178} {\path{doi:10.1111/cgf.12178}}.
\newline\urlprefix\url{https://onlinelibrary.wiley.com/doi/10.1111/cgf.12178}

\bibitem{luo2001}
{Bin Luo}, E.~Hancock, \href{http://ieeexplore.ieee.org/document/954602/}{{Structural graph matching using the EM algorithm and singular value decomposition}}, IEEE Transactions on Pattern Analysis and Machine Intelligence 23~(10) (2001) 1120--1136.
\newblock \href {https://doi.org/10.1109/34.954602} {\path{doi:10.1109/34.954602}}.
\newline\urlprefix\url{http://ieeexplore.ieee.org/document/954602/}

\bibitem{Tombari2010}
F.~Tombari, S.~Salti, L.~{Di Stefano}, \href{http://link.springer.com/10.1007/978-3-642-15558-1_26}{{Unique Signatures of Histograms for Local Surface Description}}, 2010, pp. 356--369.
\newblock \href {https://doi.org/10.1007/978-3-642-15558-1_26} {\path{doi:10.1007/978-3-642-15558-1_26}}.
\newline\urlprefix\url{http://link.springer.com/10.1007/978-3-642-15558-1_26}

\bibitem{Drost2012}
B.~Drost, S.~Ilic, \href{http://ieeexplore.ieee.org/document/6374971/}{{3D Object Detection and Localization Using Multimodal Point Pair Features}}, in: 2012 Second International Conference on 3D Imaging, Modeling, Processing, Visualization \& Transmission, IEEE, 2012, pp. 9--16.
\newblock \href {https://doi.org/10.1109/3DIMPVT.2012.53} {\path{doi:10.1109/3DIMPVT.2012.53}}.
\newline\urlprefix\url{http://ieeexplore.ieee.org/document/6374971/}

\bibitem{Rusu2009}
R.~B. Rusu, N.~Blodow, M.~Beetz, \href{http://ieeexplore.ieee.org/document/5152473/}{{Fast Point Feature Histograms (FPFH) for 3D registration}}, in: 2009 IEEE International Conference on Robotics and Automation, IEEE, 2009, pp. 3212--3217.
\newblock \href {https://doi.org/10.1109/ROBOT.2009.5152473} {\path{doi:10.1109/ROBOT.2009.5152473}}.
\newline\urlprefix\url{http://ieeexplore.ieee.org/document/5152473/}

\bibitem{Qi2017}
C.~R. Qi, L.~Yi, H.~Su, L.~J. Guibas, \href{http://arxiv.org/abs/1706.02413}{{PointNet++: Deep Hierarchical Feature Learning on Point Sets in a Metric Space}}, in: 30th Neural Information Processing Systems (NIPS 2017), 2017.
\newblock \href {http://arxiv.org/abs/1706.02413} {\path{arXiv:1706.02413}}.
\newline\urlprefix\url{http://arxiv.org/abs/1706.02413}

\bibitem{deng2018}
H.~Deng, T.~Birdal, S.~Ilic, \href{http://link.springer.com/10.1007/978-3-030-01228-1_37 https://link.springer.com/10.1007/978-3-030-01228-1_37}{{PPF-FoldNet: Unsupervised Learning of Rotation Invariant 3D Local Descriptors}}, in: ECCV 2018. Lecture Notes in Computer Science, 2018, pp. 620--638.
\newblock \href {https://doi.org/10.1007/978-3-030-01228-1_37} {\path{doi:10.1007/978-3-030-01228-1_37}}.
\newline\urlprefix\url{http://link.springer.com/10.1007/978-3-030-01228-1_37 https://link.springer.com/10.1007/978-3-030-01228-1_37}

\bibitem{Deng2018a}
H.~Deng, T.~Birdal, S.~Ilic, \href{https://ieeexplore.ieee.org/document/8578126/}{{PPFNet: Global Context Aware Local Features for Robust 3D Point Matching}}, in: 2018 IEEE/CVF Conference on Computer Vision and Pattern Recognition, IEEE, 2018, pp. 195--205.
\newblock \href {https://doi.org/10.1109/CVPR.2018.00028} {\path{doi:10.1109/CVPR.2018.00028}}.
\newline\urlprefix\url{https://ieeexplore.ieee.org/document/8578126/}

\bibitem{choy2019}
C.~Choy, J.~Park, V.~Koltun, \href{https://ieeexplore.ieee.org/document/9009829/}{{Fully Convolutional Geometric Features}}, in: 2019 IEEE/CVF International Conference on Computer Vision (ICCV), IEEE, 2019, pp. 8957--8965.
\newblock \href {https://doi.org/10.1109/ICCV.2019.00905} {\path{doi:10.1109/ICCV.2019.00905}}.
\newline\urlprefix\url{https://ieeexplore.ieee.org/document/9009829/}

\bibitem{Bai2021}
X.~Bai, Z.~Luo, L.~Zhou, H.~Chen, L.~Li, Z.~Hu, H.~Fu, C.-L. Tai, \href{https://ieeexplore.ieee.org/document/9578333/}{{PointDSC: Robust Point Cloud Registration using Deep Spatial Consistency}}, in: 2021 IEEE/CVF Conference on Computer Vision and Pattern Recognition (CVPR), IEEE, 2021, pp. 15854--15864.
\newblock \href {https://doi.org/10.1109/CVPR46437.2021.01560} {\path{doi:10.1109/CVPR46437.2021.01560}}.
\newline\urlprefix\url{https://ieeexplore.ieee.org/document/9578333/}

\bibitem{Wei2023}
T.~Wei, Y.~Patel, A.~Shekhovtsov, J.~Matas, D.~Barath, \href{https://ieeexplore.ieee.org/document/10378404/}{{Generalized Differentiable RANSAC}}, in: 2023 IEEE/CVF International Conference on Computer Vision (ICCV), IEEE, 2023, pp. 17603--17614.
\newblock \href {https://doi.org/10.1109/ICCV51070.2023.01618} {\path{doi:10.1109/ICCV51070.2023.01618}}.
\newline\urlprefix\url{https://ieeexplore.ieee.org/document/10378404/}

\bibitem{Jiang2022}
X.~Jiang, Y.~Wang, A.~Fan, J.~Ma, \href{https://linkinghub.elsevier.com/retrieve/pii/S0924271622001666}{{Learning for mismatch removal via graph attention networks}}, ISPRS Journal of Photogrammetry and Remote Sensing 190 (2022) 181--195.
\newblock \href {https://doi.org/10.1016/j.isprsjprs.2022.06.009} {\path{doi:10.1016/j.isprsjprs.2022.06.009}}.
\newline\urlprefix\url{https://linkinghub.elsevier.com/retrieve/pii/S0924271622001666}

\bibitem{Pais2020}
G.~D. Pais, S.~Ramalingam, V.~M. Govindu, J.~C. Nascimento, R.~Chellappa, P.~Miraldo, \href{https://ieeexplore.ieee.org/document/9156303/}{{3DRegNet: A Deep Neural Network for 3D Point Registration}}, in: 2020 IEEE/CVF Conference on Computer Vision and Pattern Recognition (CVPR), IEEE, 2020, pp. 7191--7201.
\newblock \href {https://doi.org/10.1109/CVPR42600.2020.00722} {\path{doi:10.1109/CVPR42600.2020.00722}}.
\newline\urlprefix\url{https://ieeexplore.ieee.org/document/9156303/}

\bibitem{Wang2019a}
Y.~Wang, J.~Solomon, \href{https://ieeexplore.ieee.org/document/9009466/}{{Deep Closest Point: Learning Representations for Point Cloud Registration}}, in: 2019 IEEE/CVF International Conference on Computer Vision (ICCV), IEEE, 2019, pp. 3522--3531.
\newblock \href {https://doi.org/10.1109/ICCV.2019.00362} {\path{doi:10.1109/ICCV.2019.00362}}.
\newline\urlprefix\url{https://ieeexplore.ieee.org/document/9009466/}

\bibitem{Wang2019b}
Y.~Wang, J.~M. Solomon, \href{https://proceedings.neurips.cc/paper/2019/file/ebad33b3c9fa1d10327bb55f9e79e2f3-Paper.pdf}{{PRNet: Self-Supervised Learning for Partial-to-Partial Registration}}, in: H.~Wallach, H.~Larochelle, A.~Beygelzimer, F.~d\textquotesingle Alch{\'{e}}-Buc, E.~Fox, R.~Garnett (Eds.), Advances in Neural Information Processing Systems (NeurIPS 2019), Vol.~32, Curran Associates, Inc., 2019.
\newline\urlprefix\url{https://proceedings.neurips.cc/paper/2019/file/ebad33b3c9fa1d10327bb55f9e79e2f3-Paper.pdf}

\bibitem{Fu2021}
K.~Fu, S.~Liu, X.~Luo, M.~Wang, \href{https://ieeexplore.ieee.org/document/9578566/}{{Robust Point Cloud Registration Framework Based on Deep Graph Matching}}, in: 2021 IEEE/CVF Conference on Computer Vision and Pattern Recognition (CVPR), IEEE, 2021, pp. 8889--8898.
\newblock \href {https://doi.org/10.1109/CVPR46437.2021.00878} {\path{doi:10.1109/CVPR46437.2021.00878}}.
\newline\urlprefix\url{https://ieeexplore.ieee.org/document/9578566/}

\bibitem{Fu2022}
K.~Fu, J.~Luo, X.~Luo, S.~Liu, C.~Zhang, M.~Wang, \href{https://ieeexplore.ieee.org/document/9878213/}{{Robust Point Cloud Registration Framework Based on Deep Graph Matching}}, IEEE Transactions on Pattern Analysis and Machine Intelligence (2022) 1--13\href {https://doi.org/10.1109/TPAMI.2022.3204713} {\path{doi:10.1109/TPAMI.2022.3204713}}.
\newline\urlprefix\url{https://ieeexplore.ieee.org/document/9878213/}

\bibitem{Zanfir2018}
A.~Zanfir, C.~Sminchisescu, {Deep Learning of Graph Matching}, in: 2018 IEEE/CVF Conference on Computer Vision and Pattern Recognition, IEEE, 2018, pp. 2684--2693.
\newblock \href {https://doi.org/10.1109/CVPR.2018.00284} {\path{doi:10.1109/CVPR.2018.00284}}.

\bibitem{Huang2021}
S.~Huang, Z.~Gojcic, M.~Usvyatsov, A.~Wieser, K.~Schindler, \href{https://ieeexplore.ieee.org/document/9577334/}{{PREDATOR: Registration of 3D Point Clouds with Low Overlap}}, in: 2021 IEEE/CVF Conference on Computer Vision and Pattern Recognition (CVPR), IEEE, 2021, pp. 4265--4274.
\newblock \href {https://doi.org/10.1109/CVPR46437.2021.00425} {\path{doi:10.1109/CVPR46437.2021.00425}}.
\newline\urlprefix\url{https://ieeexplore.ieee.org/document/9577334/}

\bibitem{Yew2020}
Z.~J. Yew, G.~H. Lee, \href{https://ieeexplore.ieee.org/document/9157132/}{{RPM-Net: Robust Point Matching Using Learned Features}}, in: 2020 IEEE/CVF Conference on Computer Vision and Pattern Recognition (CVPR), IEEE, 2020, pp. 11821--11830.
\newblock \href {https://doi.org/10.1109/CVPR42600.2020.01184} {\path{doi:10.1109/CVPR42600.2020.01184}}.
\newline\urlprefix\url{https://ieeexplore.ieee.org/document/9157132/}

\bibitem{Yang2011}
J.~Yang, \href{https://linkinghub.elsevier.com/retrieve/pii/S0167865511000304}{{The thin plate spline robust point matching (TPS-RPM) algorithm: A revisit}}, Pattern Recognition Letters 32~(7) (2011) 910--918.
\newblock \href {https://doi.org/10.1016/j.patrec.2011.01.015} {\path{doi:10.1016/j.patrec.2011.01.015}}.
\newline\urlprefix\url{https://linkinghub.elsevier.com/retrieve/pii/S0167865511000304}

\bibitem{Xu2021}
H.~Xu, S.~Liu, G.~Wang, G.~Liu, B.~Zeng, \href{https://ieeexplore.ieee.org/document/9709963/}{{OMNet: Learning Overlapping Mask for Partial-to-Partial Point Cloud Registration}}, in: 2021 IEEE/CVF International Conference on Computer Vision (ICCV), IEEE, 2021, pp. 3112--3121.
\newblock \href {https://doi.org/10.1109/ICCV48922.2021.00312} {\path{doi:10.1109/ICCV48922.2021.00312}}.
\newline\urlprefix\url{https://ieeexplore.ieee.org/document/9709963/}

\bibitem{Qin2023}
Z.~Qin, H.~Yu, C.~Wang, Y.~Guo, Y.~Peng, S.~Ilic, D.~Hu, K.~Xu, \href{https://ieeexplore.ieee.org/document/10076895/}{{GeoTransformer: Fast and Robust Point Cloud Registration With Geometric Transformer}}, IEEE Transactions on Pattern Analysis and Machine Intelligence 45~(8) (2023) 9806--9821.
\newblock \href {https://doi.org/10.1109/TPAMI.2023.3259038} {\path{doi:10.1109/TPAMI.2023.3259038}}.
\newline\urlprefix\url{https://ieeexplore.ieee.org/document/10076895/}

\bibitem{Yuan2024}
Y.~Yuan, Y.~Wu, X.~Fan, M.~Gong, W.~Ma, Q.~Miao, \href{https://ieeexplore.ieee.org/document/10319695/}{{EGST: Enhanced Geometric Structure Transformer for Point Cloud Registration}}, IEEE Transactions on Visualization and Computer Graphics (2024) 1--13\href {https://doi.org/10.1109/TVCG.2023.3329578} {\path{doi:10.1109/TVCG.2023.3329578}}.
\newline\urlprefix\url{https://ieeexplore.ieee.org/document/10319695/}

\bibitem{Yew2022}
Z.~J. Yew, G.~H. Lee, \href{https://ieeexplore.ieee.org/document/9880077/}{{REGTR: End-to-end Point Cloud Correspondences with Transformers}}, in: 2022 IEEE/CVF Conference on Computer Vision and Pattern Recognition (CVPR), IEEE, 2022, pp. 6667--6676.
\newblock \href {https://doi.org/10.1109/CVPR52688.2022.00656} {\path{doi:10.1109/CVPR52688.2022.00656}}.
\newline\urlprefix\url{https://ieeexplore.ieee.org/document/9880077/}

\bibitem{Cuevas-Velasquez2021}
H.~Cuevas-Velasquez, A.~J. Gallego, R.~B. Fisher, \href{http://arxiv.org/abs/2111.00231}{{Two Heads are Better than One: Geometric-Latent Attention for Point Cloud Classification and Segmentation}} (oct 2021).
\newblock \href {http://arxiv.org/abs/2111.00231} {\path{arXiv:2111.00231}}.
\newline\urlprefix\url{http://arxiv.org/abs/2111.00231}

\bibitem{lin2020}
Y.~Lin, Z.~Yan, H.~Huang, D.~Du, L.~Liu, S.~Cui, X.~Han, \href{https://ieeexplore.ieee.org/document/9157670/}{{FPConv: Learning Local Flattening for Point Convolution}}, in: 2020 IEEE/CVF Conference on Computer Vision and Pattern Recognition (CVPR), IEEE, 2020, pp. 4292--4301.
\newblock \href {https://doi.org/10.1109/CVPR42600.2020.00435} {\path{doi:10.1109/CVPR42600.2020.00435}}.
\newline\urlprefix\url{https://ieeexplore.ieee.org/document/9157670/}

\bibitem{paszke2017}
A.~Paszke, S.~Gross, S.~Chintala, G.~Chanan, E.~Yang, Z.~DeVito, Z.~Lin, A.~Desmaison, L.~Antiga, A.~Lerer, {Automatic differentiation in pytorch}, in: NIPS 2017 Autodiff Workshop: The Future of Gradient-based Machine Learning Software and Techniques, Long Beach, CA, US, 2017.

\bibitem{wu2015}
{Zhirong Wu}, S.~Song, A.~Khosla, {Fisher Yu}, {Linguang Zhang}, {Xiaoou Tang}, J.~Xiao, \href{https://ieeexplore.ieee.org/document/7298801/}{{3D ShapeNets: A deep representation for volumetric shapes}}, in: 2015 IEEE Conference on Computer Vision and Pattern Recognition (CVPR), IEEE, 2015, pp. 1912--1920.
\newblock \href {https://doi.org/10.1109/CVPR.2015.7298801} {\path{doi:10.1109/CVPR.2015.7298801}}.
\newline\urlprefix\url{https://ieeexplore.ieee.org/document/7298801/}

\end{thebibliography}

\end{document}